\pdfoutput=1

\documentclass[11pt]{article}
\usepackage[final]{acl}

\usepackage{times}
\usepackage{latexsym}
\usepackage[T1]{fontenc}
\usepackage[utf8]{inputenc}
\usepackage{microtype}
\usepackage{inconsolata}
\usepackage{graphicx}
\usepackage{hyperref}
\usepackage{url}
\usepackage{booktabs}
\usepackage{amsmath}
\usepackage{colortbl}
\usepackage{cleveref}
\usepackage{placeins}
\usepackage{tabularx}
\usepackage{arydshln}
\usepackage{makecell}
\usepackage{multicol}
\usepackage{multirow}

\makeatletter
\newcommand*{\rom}[1]{\expandafter\@slowromancap\romannumeral #1@}
\makeatother

\usepackage{CJKutf8}
\newcommand{\chinese}[1]{{\begin{CJK*}{UTF8}{gkai} #1 \end{CJK*}}}

\definecolor{bblue}{HTML}{648FFF}
\definecolor{rred}{HTML}{DC267F}
\definecolor{yyellow}{HTML}{FFB000}
\DeclareRobustCommand{\hlred}[1]{{\textcolor{rred}{#1}}}
\DeclareRobustCommand{\hlblue}[1]{{\textcolor{bblue}{#1}}}
\DeclareRobustCommand{\hlyellow}[1]{{\textcolor{yyellow}{#1}}}

\title{Fine-Tuning Large Language Models to Translate: \\ Will a Touch of Noisy Data in Misaligned Languages Suffice?}

\author{
Dawei Zhu$^{1}$\qquad
Pinzhen Chen$^{2}$\qquad
Miaoran Zhang$^{1}$\\
\textbf{Barry Haddow$^{2}$\qquad
Xiaoyu Shen$^{3}$\thanks{Corresponding author (xyshen@eitech.edu.cn)}\qquad
Dietrich Klakow$^{1}$
}\\
$^1$Saarland University, Saarland Informatics Campus\qquad$^2$University of Edinburgh\\
$^3$Digital Twin Institute, Eastern Institute of Technology, Ningbo\\
\texttt{\{dzhu,mzhang\}@lsv.uni-saarland.de \qquad pinzhen.chen@ed.ac.uk}
}

\begin{document}
\maketitle
\begin{abstract}
Traditionally, success in multilingual machine translation can be attributed to three key factors in training data: large volume, diverse translation directions, and high quality. In the current practice of fine-tuning large language models (LLMs) for translation, we revisit the importance of these factors. We find that LLMs display strong translation capability after being fine-tuned on as few as 32 parallel sentences and that fine-tuning on a single translation direction enables translation in multiple directions. However, the choice of direction is critical: fine-tuning LLMs with only English on the target side can lead to task misinterpretation, which hinders translation into non-English languages. Problems also arise when noisy synthetic data is placed on the target side, especially when the target language is well-represented in LLM pre-training. Yet interestingly, synthesized data in an under-represented language has a less pronounced effect. Our findings suggest that when adapting LLMs to translation, the requirement on data quantity can be eased but careful considerations are still crucial to prevent an LLM from exploiting unintended data biases.\footnote{Code available at: \href{https://github.com/uds-lsv/mt-sft}{github.com/uds-lsv/mt-sft}.}
\end{abstract}

\section{Introduction}

Large language models (LLMs) have reached new heights in various NLP tasks \citep{radford2019language, Brown2020language, touvron2023llama2, jiang2023mistral}. Supervised fine-tuning \citep[SFT,][alternatively, instruction tuning or simply fine-tuning in some literature]{ouyang2022_instructgpt} further prepares these models for better generalization and reliability in downstream tasks by training on task input-output data combined with instructions in natural languages \citep{sanh2022multitask, wei2022finetuned, mishra-etal-2022-cross}. In this research direction, various works have studied the ``scaling up'' of SFT data size, number of languages, etc \citep{chung2022scaling, muennighoff-etal-2023-crosslingual}. On the other hand, recent papers also embraced the philosophy of ``less is more'' by achieving strong results with a small set of high-quality training instances, claiming a ``superficial alignment hypothesis'' \citep{zhou2023lima} with similar findings by others.

This work investigates the role of SFT \textit{data} in aligning LLMs to machine translation (MT), a cross-lingual generation task with high demands in practical domains. Prior research has found fine-tuning to improve translation performance~\citep{zhang-etal-2023-machine} and more recent works also integrated continued pre-training with more data to provide further improvement~\citep{xu2024alma,alves2024tower}. For encoder-decoder models, \citet{wu-etal-2024-far} used little data to enable an English-centric model to translate between any two languages. Nonetheless, the feasibility of ``less is more'' in LLM translation fine-tuning is rather under-explored. In translation prompting, researchers have suggested that a model's translation capability can be attributed to the bilingual signals exposed during pre-training \citep{briakou-etal-2023-searching} and task recognition in LLM layers \citep{Sia2024WhereDI}, hinting that the translation capability has been picked up during pre-training. A natural question follows: \textit{Can we put reduced effort into data?}

From a data efficiency perspective, we squeeze the translation SFT data to a mere size of 32 or the translation direction to 1 for multilingual translation, for which we believe LLMs already possess a strong pre-trained foundation in multilingual understanding and generation. Beyond quantity and language diversity, we perform SFT on synthesized data via machine translation, which is a common data augmentation practice for under-served languages. To summarize, our analysis is grounded in the task of MT, with ``scaling down'' in mind. In multiple dimensions---data size (\S\ref{sec:size_exp}), translation direction (\S\ref{sec:direction_exp}~and~\S\ref{sec:exp_unseen_lanuages}), and data synthesis (\S\ref{sec:quality_exp})---our findings verify, complement, and refine the existing superficial alignment hypothesis for fine-tuning LLMs for translation tasks:
\begin{enumerate}
    \item 32 data instances successfully enable an LLM to translate in 11 directions. More data still helps but the return diminishes.

    \item Data in a single translation direction can effectively align an LLM to translate to and from multiple directions. Yet, it is crucial to pick the right direction---we recommend not placing English on the target side.

    \item When fine-tuning on lower-quality synthetic data, LLMs are affected if the data is placed on the target side, but they show greater resilience against such flaws in low-resource languages, which are less represented during pre-training.
\end{enumerate}

\section{Preliminaries}
\subsection{Supervised fine-tuning}
In this work, we perform SFT to prepare pre-trained LLMs for MT. Let $S$ denote a source input and $T=[t_1,t_2,...,t_{|T|}]$ denote a target-side reference. We start with placing the input into a prompt template by applying $\mathcal{I}(\cdot)$ to $S$. For each training instance, the instruction template is randomly selected from a pre-defined pool. We fine-tune an LLM parameterized by $\theta$ by optimizing the log-likelihood:

\vspace{-1ex}
\begin{small}
\begin{align}
    \nonumber\mathcal{L}_{SFT}(\mathcal{I}(S), T;\theta)&=-\log P(T|\mathcal{I}(S); \theta) \\
    \nonumber&=-\log\prod_{k=1}^{|T|} P(t_{k}|t_{<k},\mathcal{I}(S);\theta) \\
    \nonumber&=-\sum_{k=1}^{|T|} \log P(t_{k}|t_{<k},\mathcal{I}(S);\theta) 
\end{align}
\end{small}

\subsection{Superficial alignment hypothesis}
\citet{zhou2023lima} claim that a model’s knowledge and capabilities are acquired almost entirely during pre-training, and the effect of alignment tuning might be ``superficial'', in that it teaches the model the format for interacting with users. This idea is further supported by recent works \citep{lin2024urial, ghosh2024closer}. However, to what extent this applies to multilingual translation in LLMs is little known. To bridge this gap, we conduct a series of controlled experiments on fine-tuning LLMs for translation, complementing previous research across three dimensions. First, we study the parallel data efficiency in the era of LLMs, aiming to determine the minimum data needed for effective model alignment to the translation task. Next, we explore the scope of alignment by probing whether aligning one translation direction influences other directions. Finally, we investigate how synthesized fine-tuning data quality impacts the LLMs' behaviour in generating translations.

\section{Experiments and Results}

\subsection{Experimental setup}
\label{sec:default_experimental_setup}

\paragraph{Training.} By default, we take the test sets from WMT17 to WMT20 as our parallel training data \citep{bojar-etal-2017-findings,bojar-etal-2018-findings,barrault-etal-2019-findings,barrault-etal-2020-findings}; we also use the development sets in WMT21 \citep{akhbardeh-etal-2021-findings} for training if a language pair of interest is not available in earlier years. The specific training data configurations will be detailed in the subsequent sections. The test sets from WMT21 are used for validation. Detailed data statistics can be found in \Cref{appendix:sec:dataset_info}. The LLM we use for SFT is the base version of Llama-2 7B \citep{touvron2023llama2}. When performing SFT, we use a learning rate of 5e-6, an effective batch size of 64, and a linear learning rate scheduling with a warmup ratio of 0.1. We select the model checkpoint based on COMET scores on the validation sets.\footnote{In our preliminary experiments, we found that validation perplexity has a relatively weak correlation with COMET scores measured on the validation set, similar to earlier findings \citep{ouyang2022_instructgpt}.} To form the model input for SFT, we feed the source sentence into the Alpaca prompt template \citep{Taori2023}, supplementing it with a translation instruction that is randomly selected from a pool of 31 diverse instructions. Refer to \Cref{appendix:tab:instruction_collection} in the appendix for a complete list of templates.

\begin{figure*}[ht]
    \centering
    \includegraphics[width=\linewidth]{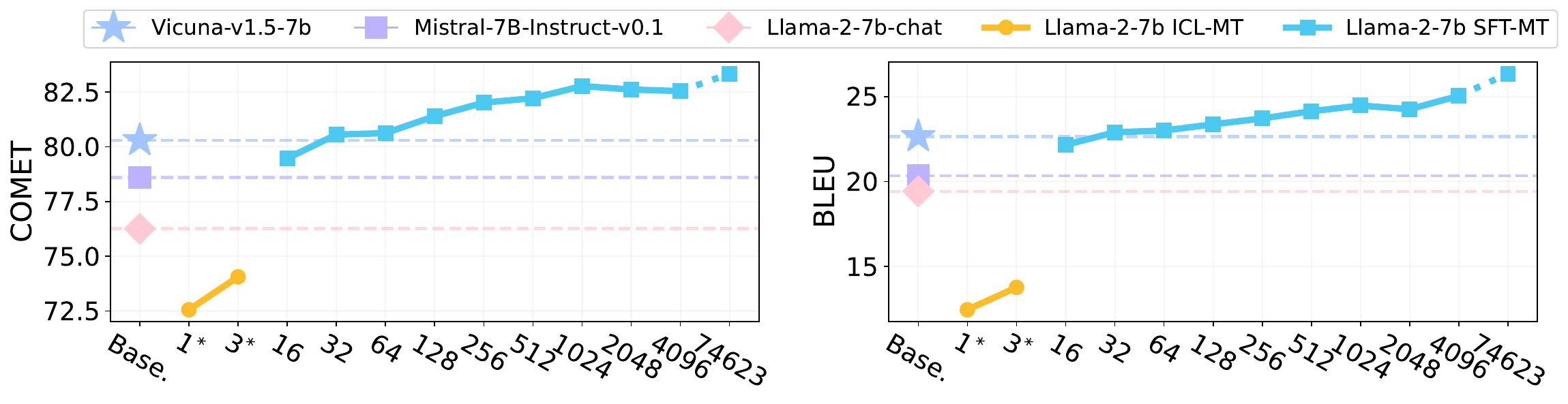}
    \caption{Performance comparison between instruction-tuned baselines and Llama-2 fine-tuned with different training data sizes. Average COMET (left) and BLEU (right) scores across 11 translation directions are presented. For training data sizes of 1 and 3, ICL is applied, marked with an asterisk ``$^*$''; otherwise, we perform SFT. With only 32 training examples for SFT, Llama-2 outperforms general-purpose, instruction-tuned baselines. Base.: instruction-tuned baseline models. See individual performance for the 11 translation directions in Appendix~\ref{appendix:sec:model_performance_sample_size_sep}.}
    \label{fig:size_scaling}
\end{figure*}

\paragraph{Evaluation.} We primarily evaluate the models on the WMT22 test sets \citep{kocmi-etal-2022-findings} covering 11 translation directions: en$\leftrightarrow$cs, en$\leftrightarrow$de, en$\leftrightarrow$jp, en$\leftrightarrow$ru, en$\leftrightarrow$zh, and en$\rightarrow$hr.\footnote{Language codes: cs=Czech, de=German, hr=Croatian, jp=Japanese, ru=Russian, zh=Chinese. ``$\leftrightarrow$'' means that both translation directions are covered. Note that only en$\rightarrow$hr is available in WMT22 but not hr$\rightarrow$en.} Languages in these 11 directions are explicitly included in Llama-2's pre-training corpus. In Section~\ref{sec:exp_unseen_lanuages}, we extend our evaluation to translation directions involving medium and low resource languages: Icelandic and Hausa (i.e., en$\leftrightarrow$is, en$\leftrightarrow$ha), which comes from WMT21's test set. At inference time, a fixed translation instruction is applied (\Cref{appendix:tab:instruction_collection}~row~1). We use beam search with a beam size of 4 for generation, as our preliminary results indicate that it offers better translation quality than sampling-based generation, an observation consistent with recent works~\cite{jiao2023_parrot, zeng2024_tim}. The maximum generation length is set to 256 tokens. We used a reference-based COMET22 checkpoint\footnote{Specifically, COMET is reported on a scale of 0 to 100 as opposed to its raw 0 to 1 range.}~\citep{rei-etal-2020-comet} and BLEU~\citep{papineni-etal-2002-bleu} as the evaluation metrics. See \Cref{sec:comet-bleu-details} for detailed software configurations.

\subsection{How much SFT data enables LLMs to translate?}
\label{sec:size_exp}

Recent works in machine translation suggest that pre-trained LLMs require significantly less parallel data for fine-tuning (via SFT), compared to training conventional translation models from scratch. However, the SFT process in these works still operates with an order of $10^5$ parallel samples~\citep[][i.a.]{jiao2023_parrot, zhang-etal-2023-machine, zeng2024_tim, xu2024alma}, without a clear justification for selecting this specific data size and source. This raises a pivotal question, inspired by the recently proposed ``superficial alignment hypothesis''~\citep{zhou2023lima}: Is SFT mainly a method for superficially aligning LLMs for translation tasks? If so, what is the actual minimal amount of data required to achieve effective ``alignment''?

\paragraph{Setup.} We fine-tune Llama-2 7B using different numbers of training samples and evaluate the multilingual translation performance of the resulting models. We collect training data covering 10 translation directions: en$\leftrightarrow$\{cs, de, jp, ru, zh\}. The training data sourced from WMT17-20 contains a total of 74,623 parallel examples. Note that the training samples across translation directions are not evenly distributed. To create training sets of varying sizes, we subsample the original data into subsets that are powers of 2, starting from 16 ($2^4$) and ending with 4096 ($2^{12}$); larger subsets always contain smaller ones. To ensure balanced language representation in our subsets, we distribute samples as evenly as possible among the language pairs.\footnote{For example, the data size distribution for our 32-example training set is $[4, 4, 3, 3, 3, 3, 3, 3, 3, 3]$.}

We refer to the fine-tuned model as \textbf{SFT-MT}. Considering LLMs can also perform translation through prompting, we compare SFT-MT with 1- and 3-shot in-context learning (ICL), denoted as \textbf{ICL-MT}. For ICL, we randomly select demonstrations from the training set in the test direction for each test sentence. We do not consider Llama-2's zero-shot performance because, although it sometimes produces acceptable translations in the beginning, it often continues generating, which makes it difficult to accurately estimate its performance. Lastly, since LLMs fine-tuned on diverse tasks also serve as strong translation systems \citep{zhu-preference-mt}, we compare our models with open-source general-purpose instruction-tuned LLMs, which we denote as \textbf{IT-LLM}. These include Vicuna-v1.5-7b \citep{vicuna2023}, Mistral-7b-Instruct \citep{jiang2023mistral}, and Llama-2-7b-chat \citep{touvron2023llama2}.\footnote{\href{https://huggingface.co/lmsys/vicuna-7b-v1.5}{lmsys/vicuna-7b-v1.5}, \href{https://huggingface.co/mistralai/Mistral-7B-Instruct-v0.1}{Mistral-7B-Instruct-v0.1}, and \href{https://huggingface.co/meta-llama/Llama-2-7b-chat-hf}{meta-llama/Llama-2-7b-chat-hf}.}

\label{study_translation_direction}
\begin{figure*}[ht]
    \centering
    \includegraphics[width=\linewidth]{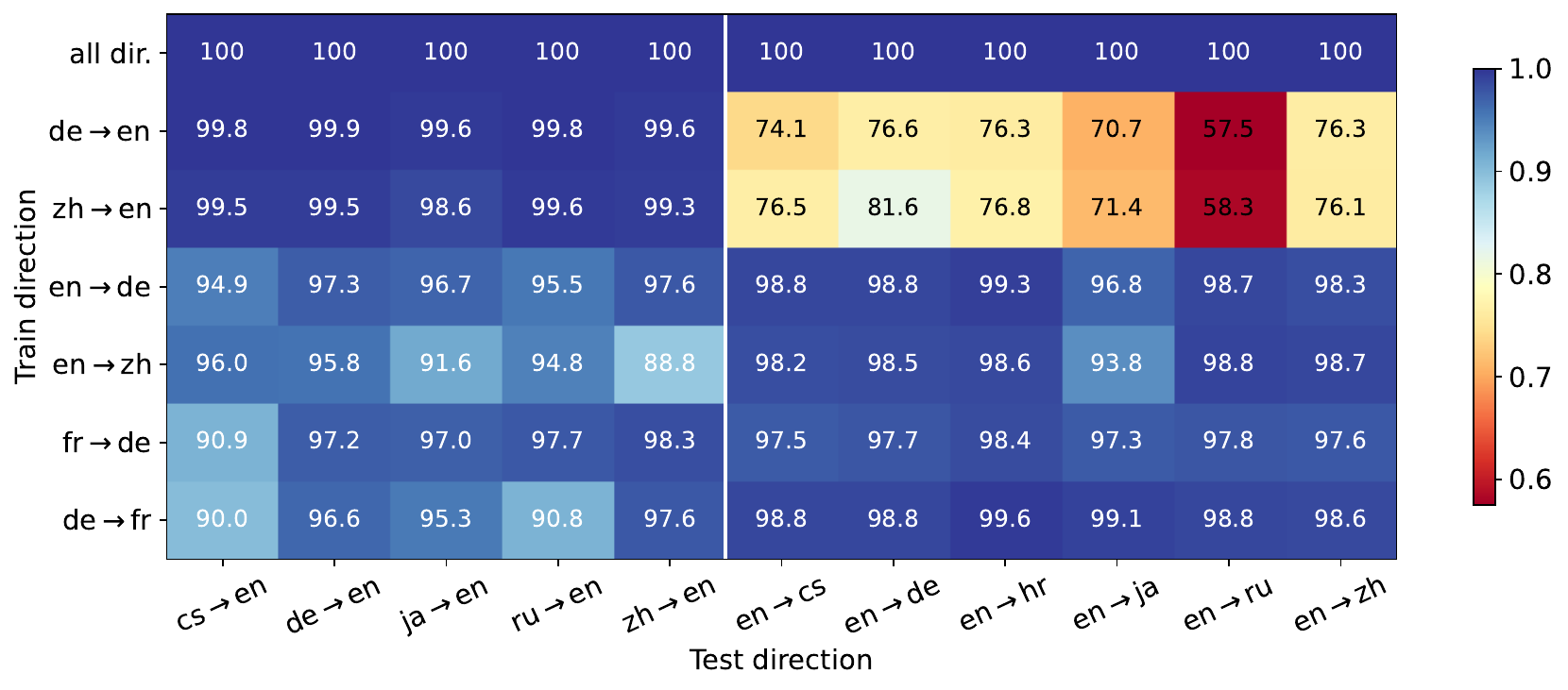}
    \caption{Normalized COMET score (as a \% of performance from fine-tuning on an equivalent sized dataset of all 10 directions) resulted from varying combinations of train and test translation directions. In most cases, Llama-2 fine-tuned on a single translation direction can effectively translate across other directions, achieving performance comparable to models trained on all directions, with a few exceptions when trained on X$\rightarrow$en but tested on en$\rightarrow$X. Performance measured in BLEU score is provided in~\Cref{appendix:sec:model_performance_vary_training_directions}.}
    \label{fig:single_direction}
\end{figure*}

\paragraph{Results.} Figure~\ref{fig:size_scaling} illustrates the effect of varying training sizes on translation performance. In both 1- and 3-shot cases, ICL-MT underperforms IT-LLM baselines like Llama-2-7b-chat despite sharing the same foundation model, indicating that a few in-context demonstrations may not effectively align Llama-2 for translation.

However, performance significantly improves when Llama-2 is fine-tuned with just 16 samples. With further increases in the training size to 32 samples, Llama-2 performs on par with or surpasses all three IT-LLM baselines in both COMET and BLEU metrics. This suggests that a handful of high-quality parallel data can effectively specialize the model into a performant translation system. Increasing parallel data further boosts performance, though with diminishing returns: the COMET score rises by an average of 2 points when expanding from 32 to 1024 samples, but only by 0.5 points when increasing further from 1024 to 75K samples (full training set). Given that it is unlikely that these 32 training samples ``teach'' Llama-2 new translation skills, this shows strong evidence that superficial alignment applies to MT. We observe a similar trend in Mistral-7B and Llama-2-13B. Refer to \Cref{appendix:sec:model_performance_sample_size_sep} for their performance across varying data sizes. In summary, effective translation alignment begins with minimal training data, revealing \textbf{less is good alignment and more is better with diminishing gains}.

\subsection{Do we need to include all directions?}
\label{sec:direction_exp}
In the preceding section, we follow the traditional practice in multilingual MT by including multiple translation directions during training. However, the observation that only a few dozen examples make Llama-2 translate well leads us to reconsider the necessity of including samples from all directions of interest. Specifically, will training on just a single translation direction be sufficient to help LLMs perform multilingual translation?

\paragraph{Setup.} We explore six training configurations, each focusing on a single translation direction: de$\rightarrow$en, zh$\rightarrow$en, en$\rightarrow$de, en$\rightarrow$zh, fr$\rightarrow$de, and de$\rightarrow$fr. These configurations include cases where English appears on the source side, the target side, as well as settings with English excluded, to investigate if specific languages have a different impact on the overall performance. The training size is set to 1024 for SFT. Evaluations are conducted across the same 11 test directions as used in the previous section. Additionally, we explore similar settings in ICL, where we present demonstrations with translation directions that do not match those used in evaluations, to determine if the mechanisms of both SFT and ICL exhibit similarities. Lastly, we conduct a joint evaluation, progressively expanding both the training size and the range of covered translation directions to understand the combined effect of these factors.

\begin{table*}[t!]
\centering
\small
\begin{minipage}{.49\linewidth}
\centering
\begin{tabular}{ccccc}
\toprule
\multicolumn{5}{c}{Evaluation on de$\rightarrow$en} \\
\midrule
\multirow[b]{2}{*}{\makecell{demo\\lang}} & \multicolumn{2}{c}{1-shot} & \multicolumn{2}{c}{3-shot} \\
\cmidrule(lr){2-3}\cmidrule(lr){4-5}
 & COMET & BLEU & COMET & BLEU \\
\midrule
de$\rightarrow$en & 73.47 & 19.7 & 75.04 & 22.4 \\
\cmidrule(lr){1-5}
en$\rightarrow$de & 55.96 & \phantom{0}7.3  & 44.39 & \phantom{0}3.5 \\
de$\rightarrow$fr & 66.35 & 12.1 & 64.61 & 17.6 \\
fr$\rightarrow$de & 58.06 & \phantom{0}7.8  & 57.13 & 10.5 \\
zh$\rightarrow$en & 56.66 & 10.7 & 54.82 & \phantom{0}7.1 \\
en$\rightarrow$zh & 51.30 & \phantom{0}7.8  & 56.87 & \phantom{0}1.8 \\
\bottomrule
\end{tabular}
\end{minipage}
\hfill
\begin{minipage}{.49\linewidth}
\centering
\begin{tabular}{ccccc}
\toprule
\multicolumn{5}{c}{Evaluation on en$\rightarrow$de} \\
\midrule
\multirow[b]{2}{*}{\makecell{demo\\lang}} & \multicolumn{2}{c}{1-shot} & \multicolumn{2}{c}{3-shot} \\
\cmidrule(lr){2-3}\cmidrule(lr){4-5}
 & COMET & BLEU & COMET & BLEU \\
\midrule
en$\rightarrow$de & 67.37 & 10.5 & 69.80 & 14.3 \\
\cmidrule(lr){1-5}
de$\rightarrow$en & 57.83 & \phantom{0}8.7  & 45.54 & \phantom{0}5.0 \\
en$\rightarrow$zh & 59.76 & \phantom{0}9.5  & 59.53 & \phantom{0}8.4 \\
zh$\rightarrow$en & 47.31 & \phantom{0}4.5  & 49.24 & \phantom{0}5.0 \\
fr$\rightarrow$de & 59.36 & \phantom{0}8.6  & 66.01 & 12.9 \\
de$\rightarrow$fr & 60.70 & 11.0 & 61.76 & 11.3 \\
\bottomrule
\end{tabular}
\end{minipage}
\caption{ICL-MT performance with aligned vs.~misaligned demonstrations, evaluated on de$\rightarrow$en and en$\rightarrow$de. 1-shot/3-shot: using 1 or 3 demonstrations randomly sampled from the training set. Misaligned demonstrations consistently cause a substantial performance drop.}
\label{tab:icl-mismatch-direction}
\end{table*}

\paragraph{SFT results.} Figure~\ref{fig:single_direction} demonstrates the normalized performance of Llama-2 when fine-tuned in various single directions. Remarkably, training with just one direction enables Llama-2 to translate between multiple languages. For instance, after fine-tuning on de$\rightarrow$en or zh$\rightarrow$en, the model can translate from all considered languages to English, scoring at least $98.6\%$ of the original COMET scores for training on all directions. Similarly, the model fine-tuned on en$\rightarrow$de, en$\rightarrow$zh, fr$\rightarrow$de or de$\rightarrow$fr also demonstrates only a slight performance decline when translating from English. 

Notable declines are observed in two scenarios: (1) trained to translate to English and evaluated on translating to non-English; and (2) trained to translate to non-English and evaluated on translating to English.\footnote{Analysis of model outputs reveals that they often merely echo the source sentence, ignoring the translation instruction.} Of these two scenarios, scenario 1 exhibits a much larger performance drop. The fact that both scenarios involve a mismatch between using English and non-English suggests that \emph{Llama-2, as an English-centric LLM, may process English differently compared to other languages}. When fine-tuned for English generation, the model may misinterpret the task as only generating in English. Generalization among non-English languages is much easier than generalization between English and non-English languages, as evidenced by the negligible performance drop when fine-tuning and testing on two vastly different language pairs such as de$\rightarrow$fr and en$\rightarrow$zh. 
Overall, the findings suggest that \textbf{SFT in one translation direction effectively enables the many directions, though avoiding misinterpretation is crucial}.

\begin{figure}[ht]
    \centering\includegraphics[width=\linewidth,trim={0ex 0ex 0ex 4ex},clip]{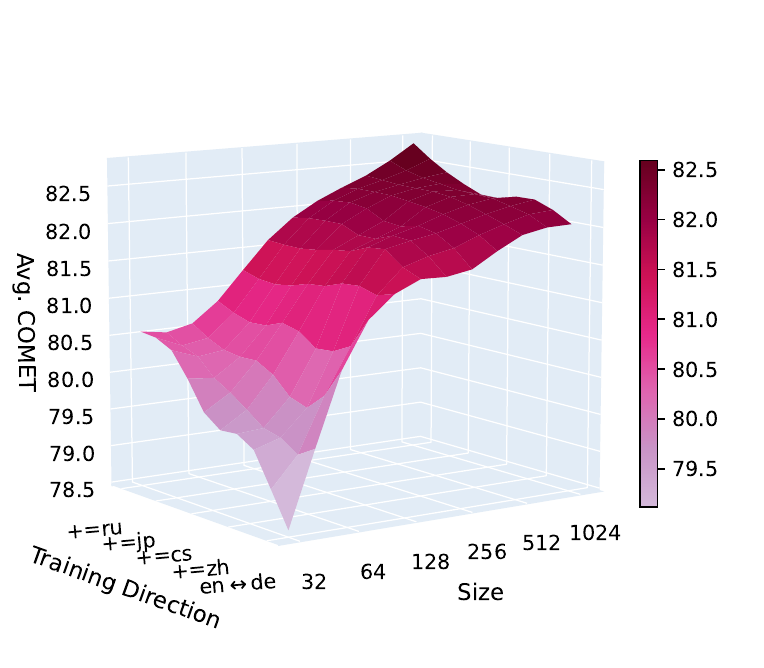}
    \caption{Average performance (in COMET) across 11 test directions for models trained with varying data sizes and directions. Both factors positively impact performance. +=: training directions added on top of previous directions; two directions are added at each time. For example, ``+=ru'' covers 10 directions: en $\leftrightarrow$ \{de, zh, cs, jp, ru\}. Performance on individual test directions is provided in~\Cref{appendix:sec:model_performance_direction_sep}.}
    \label{fig:size_direction_joint}
\end{figure}

\paragraph{ICL results.} We also provide results of performing ICL with misaligned translation directions between demonstration and test in~\Cref{tab:icl-mismatch-direction}. It can be seen that misaligned demonstrations significantly degrade translation performance, with 3-shot be often worse than 1-shot. We observe that the model may output Chinese characters, emojis, time, etc., but no clear error patterns are observed. This contrasts sharply with findings from SFT: \textbf{while SFT can recognize the format of translation, ICL requires language-aligned demonstrations.}

\begin{figure*}[ht]
    \centering
    \includegraphics[width=\linewidth]{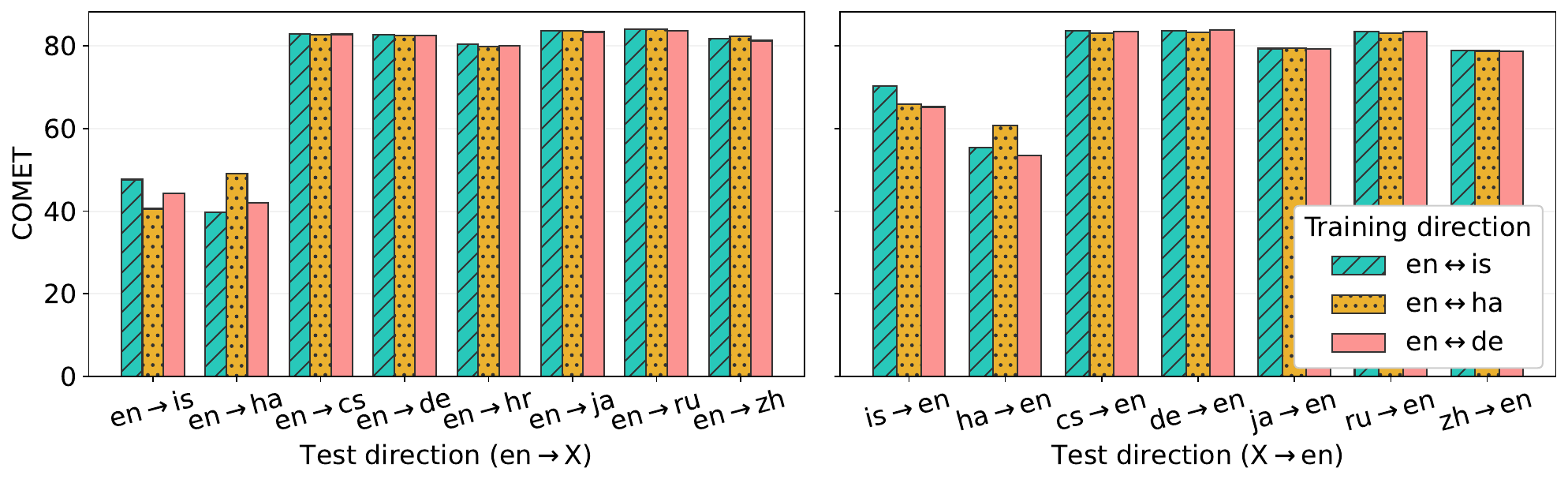}
        \caption{Model performance (in COMET) across 15 translation directions under different training configurations. Training models on \textit{unseen} languages (en$\leftrightarrow$is, en$\leftrightarrow$ha) results in slight improvements in translating these languages compared to models trained on en$\leftrightarrow$de. The differences in performance when translating between \textit{seen} languages are minimal across all training configurations. Performance measured in BLEU score is provided in Appendix~\ref{appendix:sec:model_performance_unseen_sep}.}

    \label{fig:sft_unseen_direction}
\end{figure*}

\paragraph{Joint evaluation.} \Cref{fig:size_direction_joint} presents a joint evaluation of size and translation direction. For small training sizes, covering diverse translation directions in training proves to be beneficial. However, the benefits of such diversity level off as the training size increases. With a training size of 1024, models trained exclusively on two directions, en$\leftrightarrow$de, perform on par with those trained on all directions.

\subsection{Can alignment be achieved for unseen languages?}
\label{sec:exp_unseen_lanuages}
Previous sections focus on translation directions involving languages explicitly included in Llama-2's pre-training corpus. We now extend our investigation to languages that do not have an identified presence of over 0.005\% in the pre-training data \citep[c.f.][p22]{touvron2023llama2}, referred to as \textit{unseen} languages. Here we seek answers to two questions: (1) Can we effectively make Llama-2 translate both from and to unseen languages by fine-tuning it with a small amount of data? (2) How well can this fine-tuned model translate from and to languages \textit{seen} in Llama?

\paragraph{Setup.} We consider three training configurations: en$\leftrightarrow$is, en$\leftrightarrow$ha, and en$\leftrightarrow$de, with Icelandic (is) and Hausa (ha) being unseen languages. en$\leftrightarrow$de serves as a control to assess Llama-2’s initial translation capabilities into unseen languages without specific fine-tuning. The training size is fixed at 1024 (512 samples for each direction). The test directions include the 11 directions as before, plus en$\leftrightarrow$is and en$\leftrightarrow$ha coming from the WMT21 test.

\paragraph{Results.} The results are presented in \Cref{fig:sft_unseen_direction}. It can be seen that fine-tuning on Icelandic and Hausa enhances a model's translation quality on these languages compared to the control setup, yet the gains are modest. We observe that Llama-2 manages to produce tokens in these languages, however, the translations often largely deviate from the original meanings. This suggests that it is difficult to teach models new translation directions via SFT with limited data. Interestingly, we find fine-tuning on Icelandic or Hausa does not hinder Llama-2’s ability to translate from and to all seen languages, maintaining performance levels comparable to the control scenario with en$\leftrightarrow$de. Based on these results, we propose a complement to the superficial alignment hypothesis in MT: \textbf{LLMs may learn the essence of the translation task without requiring input-output mappings in languages it ``understands'' well.}

\begin{figure*}[t]
    \centering
    \includegraphics[width=\linewidth]{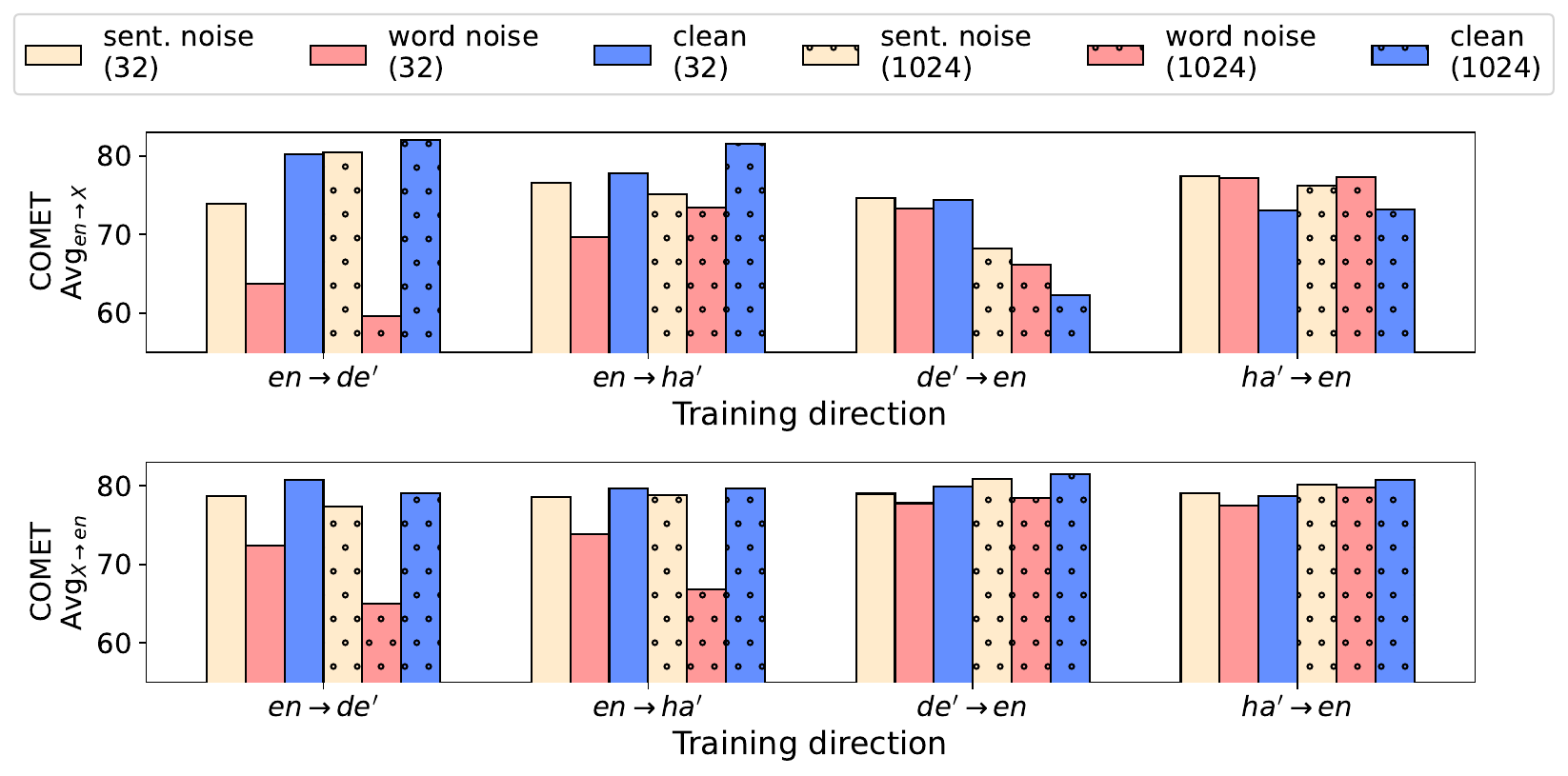}
    \caption{Model performance in COMET score varying training sizes, directions, and noise types. Top (Bottom): score averaged across all en$\rightarrow$X (X$\rightarrow$en) test directions. Training sizes considered are 32 and 1024. Generally, introducing noise on the target side tends to degrade model performance more, with the extent of impact also depending on the particular language involved. Performance measured in BLEU score is provided in Appendix~\ref{appendix:sec:model_performance_noisy_sep}.}
    \label{fig:sft_noise}
\end{figure*}

\begin{table*}[t]
    \centering
    \small
    \begin{tabular}{p{4.7cm}p{3.5cm}p{6cm}}
        \toprule
        Source & Ref./Data config. & Model output \\
        \midrule
        Das finde ich ehrlich gesagt& reference & That really bothers me, I must say. \\
         sehr ärgerlich. & literal &  The find I honest said very annoying. \\
         & en$\rightarrow$de \hlblue{clean} & I find that really annoying.\\
         & en$\rightarrow$de \hlyellow{sent. noise} & I find that honestly very annoying.\\
         & en$\rightarrow$de \hlred{word noise} & The find I honestly said very annoying. \\
         \midrule
        \chinese{以免再次发生这样的事情} & reference & So that such a thing won’t happen again. \\
                & literal & in order to avoid again happen such thing. \\
         & en$\rightarrow$de \hlblue{clean} & Let's not let it happen again. \\
         & en$\rightarrow$de \hlyellow{sent. noise} & In order not to happen again.\\
         & en$\rightarrow$de \hlred{word noise} & Avoid again happen this way. \\
        \bottomrule
    \end{tabular}
    \caption{Examples of testing Llama-2 trained on en$\rightarrow$de with 1024 clean and noisy target sentences. The test directions are de$\rightarrow$en (Top) and zh$\rightarrow$en (Bottom). The reference translation is provided by the WMT22 test set. Word-to-word references were created by the authors in consultation with native speakers. Word-level noise makes Llama-2 degenerate into a literal translator.}
    \label{tab:noise-model-ouput}
\end{table*}

\subsection{Can we use synthesized data?}
\label{sec:quality_exp}
We have observed that LLMs quickly recognize the translation task with minimal high-quality, manually curated data, but what if the quality of the training data is subpar? This situation may occur, for example when parallel data is web-crawled or machine-generated. Can LLMs still adapt to the translation task or will they overfit to the imperfections in lower-quality data, leading to degraded translation performance?

\paragraph{Setup.} We replace either the source or target sentences in the original training set with lower-quality synthesized ones. We try two types of data synthesis: one by translating entire sentences on the other side and another by concatenating word-to-word translations. Pleasingly, these correspond to back-translation \citep{sennrich-etal-2016-improving} using translation engines or bilingual word dictionaries which are practical at different levels of resource availability.  Specifically, we use the OPUS-MT suite~\citep{tiedemann-thottingal-2020-opus} to translate from English to a target non-English language.\footnote{E.g.~for de$\rightarrow$en, the process is run in en$\rightarrow$de with the created data reversed, hence the translated content is on the source side. Checkpoints are available on Hugging Face: \texttt{Helsinki-NLP/opus-mt-en-\${\{trg\}}}.} For word-level translation, we translate each space-delimited source word by feeding it into the MT model one at a time. Naturally, the synthesized versions introduce translation errors, adding ``noise'' to the training process. We investigate the impact of such noise in four translation directions: en$\rightarrow$de$'$, de$'$$\rightarrow$en, en$\rightarrow$ha$'$, and ha$'$$\rightarrow$en, where the prime ($'$) notation denotes the side that is created using translation (noised). We consider two training sizes: 32 and 1024. In this section, our evaluation focuses on the 11 translation directions described in~\Cref{sec:default_experimental_setup}. Note that although Hausa is included in the current training setup, translation directions involving Hausa are excluded from our evaluation---because performance is sub-par for unseen languages as demonstrated in~\Cref{sec:exp_unseen_lanuages}.

\paragraph{Results.} According to~\Cref{fig:sft_noise}, it can be seen that both types of data synthesis generally cause a drop in performance. However, The degree of degradation significantly varies depending on whether the noise appears on the source or target side of the translation as well as the language. Specifically, when noise is introduced to the target side, models fine-tuned on en$\rightarrow$de$'$ and en$\rightarrow$ha$'$ translations exhibit a sharp decline in performance. The impact of word noise is more severe than that of sentence noise. In the case of en$\rightarrow$de$'$, word-level synthesis causes the model to largely degenerate, leading to literal translations across many test cases across translation directions. An example of this behaviour is presented in~\Cref{tab:noise-model-ouput}. In contrast, the performance drop caused by word noise is less pronounced with en$\rightarrow$ha$'$, particularly when evaluated on en$\rightarrow$X.

Conversely, when noise is introduced on the source side, the negative impact is much smaller, and the disparity in performance degradation between the two types of noise diminishes. Even more strikingly, when evaluated on en$\rightarrow$X, having noise at the source side often outperforms the clean settings. Notably, in ~\Cref{sec:direction_exp}, we show that fine-tuning models purely on X$\rightarrow$en risks task misinterpretation, leading to low performance on en$\rightarrow$X. However, adding noise appears to mitigate this issue, resulting in improvements in both COMET and BLEU scores, especially for the ha$'\rightarrow$en case.

Summarizing the observations, Llama-2 is much more robust against the noise introduced in Hausa, likely because it has limited familiarity with the language, making it more difficult to detect and imitate imperfections present in the training data. As a result, Llama-2 tends to just recognize the essence of the translation task instead of overfitting to the biases present in low-quality data. In contrast, with German, Llama-2's understanding leads to a misinterpretation of the training objectives, such as fitting the word-level noise with a directive for literal translations. Overall, \textbf{LLMs may quickly fit translation imperfections in the training data, especially for seen languages; the resulting performance drop may be observable with just 32 training samples.}

\section{Related Work}
\subsection{What does LLM SFT bring us?}
Foundational language models become more robust and follow instructions better after being fine-tuned on task-oriented supervised data formulated as natural language text \citep{mishra-etal-2022-cross,sanh2022multitask,wei2022finetuned}. We observe diverging trends in research on instruction tuning nowadays: (1) Many works attempt to scale up instruction data in terms of the number of tasks, languages, data size, and thus implicitly increasing training updates \citep{chung2022scaling,muennighoff-etal-2023-crosslingual,wu2024laminilm,li2023bactrianx,ustun2024aya,zhang2024when}. (2) Another stream of papers, argue that instruction tuning mainly alters a base model's response style but not content or knowledge---data quality and diversity outweigh quantity \citep{zhou2023lima,mitchell2023emulator,lin2024urial,chen2024alpagasus}. This work is a continued exploration of the latter, focusing on the machine translation task. We verify the effect of size variations and include two new factors---language directions and quality---aiming to provide practical and cost-effective guidance on this matter. 

Specifically, language transfer has been demonstrated in smaller pre-trained models before LLMs \citep{wu-dredze-2019-beto,artetxe-etal-2020-cross}. For (sufficiently) multilingual models, training on certain languages might still benefit other languages at the test time \citep{choenni-etal-2023-languages}. In LLM instruction tuning, recent papers revealed cross-lingual transfer and improved robustness in unseen languages via multilingual instruction tuning with a small data sample \citep{chen2023monolingual,kew2023turning,shaham2024multilingual}. Furthermore, it has been claimed that even monolingual instruction tuning is sufficient to elicit multilingual responses in the correct languages with a key ingredient being the right learning rate \citep{chirkova2024key,chirkova2024zeroshot}. In relation to our experiments, language transfer to unseen languages might account for improved performance in language directions that are not directly fine-tuned.

\subsection{How can we use LLMs for translation?}
In the field of machine translation, earlier works provided analysis of general-purpose prompting \citep{vilar-etal-2023-prompting,agrawal-etal-2023-context,zhang2023prompting} followed by a blossom of strategies focusing on specific aspects of the translation process \citep{sarti-etal-2023-ramp,ghazvininejad2023dictionarybased,he2023exploring,moslem-etal-2023-adaptive,chen2023iterative,raunak-etal-2023-leveraging}. Nonetheless, as shown in our experimental results, few-shot prompting is not on par with using instruction-tuned models, illustrating the importance of further understanding the role of instruction tuning in translation tasks.

In terms of fine-tuning LLMs for translation, previous works have explored a wide range of sub-tasks: disambiguation, low-resource, document-level, and adaptive translation, etc \citep{li2023eliciting,zhang2023bayling,alves-etal-2023-steering,iyer-etal-2023-towards,mao2024tuning,wu2024adapting}. These works focus on improving translation performance and specific applications. \citet{Stap2024} show that while fine-tuning improves translation quality, it can degrade certain key LLMs' advantages, such as the contextualization ability on document-level input. Some recent research aims to enhance the translation capabilities of LLMs by incorporating human preference data \citep{jiao2023_parrot, zeng2024_tim, zhu-preference-mt} or by extending the pre-training phase before fine-tuning \citep{xu2024alma,xu2024contrastive,alves2024tower}, yet these approaches require significantly more data or computing resources. The aim of this paper is not to pursue the state of the art but to investigate the opportunities of extending instruction-tuned LLMs' translation capabilities in desirable compute-efficient scenarios. It is still worth noting that our investigation is orthogonal to previous works which employ relatively large monolingual and parallel data for continued pre-training.

\section{Conclusion and Future Work}
In this work, we conduct an in-depth analysis of fine-tuning LLMs for translation. We demonstrate that LLMs is capable of translating in multiple directions after being fine-tuned with \emph{minimal low-quality training data in a single direction}. While this suggests pre-trained LLMs inherently possess multilingual translation capabilities which only need to be unlocked by aligning with the correct task format, we discover pitfalls and lessons in aligning LLMs; while LLMs make efforts to adjust to the translation task, they are good at imitating other patterns such as the noise in the parallel data. Future work could explore robust training methods that align LLMs with translation while minimizing the risk of overfitting to low-quality data.

\section*{Limitations}
This work offers a range of insights into fine-tuning LLMs for translation. However, our study is not exhaustive and is subject to the following limitations.

\paragraph{Model size and diversity.} Throughout our systematic study, we fine-tuned Llama-2 7B, Llama-2 12B, and Mistral 7B. These are strong and feasible options when the work is carried out. It is important to verify the generalizability of our findings to models with different capabilities or of different sizes. 

\paragraph{Non-English~centric MT.} Our evaluation is English-centric, which is the condition of most LLM pre-training. Findings will be more comprehensive if future work can extend it to translation directions not involving English.

\paragraph{State-of-the-art performance.} Our research primarily explores how SFT enables LLM to translate to uncover data-efficient strategies in SFT and identify associated pitfalls. Recent studies have demonstrated that translation capabilities can be further enhanced through techniques such as continual pre-training~\citep{xu2024alma,alves2024tower} and preference learning~\citep{xu2024contrastive, zhu-preference-mt}. However, these methods require significantly more training resources, which may pose challenges when applied to large models.

\paragraph{Fine-tuning methods.} Throughout this work, we perform SFT with full-parameter updates. It is worthwhile to explore parameter-efficient methods which bring in heavier regularization to understand whether they exhibit patterns similar to those observed in our work.

\section*{Ethical considerations}
Our work's sole aim is to study the influence of data factors in applying supervised fine-tuning to large language models. We expect minimal social risks to be associated with our efforts.

\section*{Acknowledgments}

We sincerely thank the reviewers of this work for their constructive and insightful feedback.

Pinzhen Chen and Barry Haddow received funding from UK Research and Innovation (UKRI) under the UK government’s Horizon Europe funding guarantee [grant number 10052546]. Miaoran Zhang received funding from the DFG (German
Research Foundation) under project 232722074, SFB 1102. We thank EIT and IDT High Performance Computing Center for providing computational resources for this project. 

\bibliography{anthology, custom}

\begin{thebibliography}{66}
\providecommand{\natexlab}[1]{#1}

\bibitem[{Agrawal et~al.(2023)Agrawal, Zhou, Lewis, Zettlemoyer, and Ghazvininejad}]{agrawal-etal-2023-context}
Sweta Agrawal, Chunting Zhou, Mike Lewis, Luke Zettlemoyer, and Marjan Ghazvininejad. 2023.
\newblock \href {https://doi.org/10.18653/v1/2023.findings-acl.564} {In-context examples selection for machine translation}.
\newblock In \emph{Findings of the Association for Computational Linguistics: ACL 2023}.

\bibitem[{Akhbardeh et~al.(2021)Akhbardeh, Arkhangorodsky, Biesialska, Bojar, Chatterjee, Chaudhary, Costa-jussa, Espa{\~n}a-Bonet, Fan, Federmann, Freitag, Graham, Grundkiewicz, Haddow, Harter, Heafield, Homan, Huck, Amponsah-Kaakyire, Kasai, Khashabi, Knight, Kocmi, Koehn, Lourie, Monz, Morishita, Nagata, Nagesh, Nakazawa, Negri, Pal, Tapo, Turchi, Vydrin, and Zampieri}]{akhbardeh-etal-2021-findings}
Farhad Akhbardeh, Arkady Arkhangorodsky, Magdalena Biesialska, Ond{\v{r}}ej Bojar, Rajen Chatterjee, Vishrav Chaudhary, Marta~R. Costa-jussa, Cristina Espa{\~n}a-Bonet, Angela Fan, Christian Federmann, Markus Freitag, Yvette Graham, Roman Grundkiewicz, Barry Haddow, Leonie Harter, Kenneth Heafield, Christopher Homan, Matthias Huck, Kwabena Amponsah-Kaakyire, Jungo Kasai, Daniel Khashabi, Kevin Knight, Tom Kocmi, Philipp Koehn, Nicholas Lourie, Christof Monz, Makoto Morishita, Masaaki Nagata, Ajay Nagesh, Toshiaki Nakazawa, Matteo Negri, Santanu Pal, Allahsera~Auguste Tapo, Marco Turchi, Valentin Vydrin, and Marcos Zampieri. 2021.
\newblock \href {https://aclanthology.org/2021.wmt-1.1} {Findings of the 2021 conference on machine translation ({WMT}21)}.
\newblock In \emph{Proceedings of the Sixth Conference on Machine Translation}.

\bibitem[{Alves et~al.(2023)Alves, Guerreiro, Alves, Pombal, Rei, de~Souza, Colombo, and Martins}]{alves-etal-2023-steering}
Duarte~M. Alves, Nuno~M. Guerreiro, Jo{\~a}o Alves, Jos{\'e} Pombal, Ricardo Rei, Jos{\'e} de~Souza, Pierre Colombo, and Andre Martins. 2023.
\newblock \href {https://aclanthology.org/2023.findings-emnlp.744/} {Steering large language models for machine translation with finetuning and in-context learning}.
\newblock In \emph{Findings of the Association for Computational Linguistics: EMNLP 2023}.

\bibitem[{Alves et~al.(2024)Alves, Pombal, Guerreiro, Martins, Alves, Farajian, Peters, Rei, Fernandes, Agrawal et~al.}]{alves2024tower}
Duarte~M. Alves, Jos{\'e} Pombal, Nuno~M Guerreiro, Pedro~H Martins, Jo{\~a}o Alves, Amin Farajian, Ben Peters, Ricardo Rei, Patrick Fernandes, Sweta Agrawal, et~al. 2024.
\newblock \href {https://arxiv.org/abs/2402.17733} {Tower: An open multilingual large language model for translation-related tasks}.
\newblock \emph{arXiv preprint}.

\bibitem[{Artetxe et~al.(2020)Artetxe, Ruder, and Yogatama}]{artetxe-etal-2020-cross}
Mikel Artetxe, Sebastian Ruder, and Dani Yogatama. 2020.
\newblock \href {https://doi.org/10.18653/v1/2020.acl-main.421} {On the cross-lingual transferability of monolingual representations}.
\newblock In \emph{Proceedings of the 58th Annual Meeting of the Association for Computational Linguistics}.

\bibitem[{Barrault et~al.(2020)Barrault, Biesialska, Bojar, Costa-juss{\`a}, Federmann, Graham, Grundkiewicz, Haddow, Huck, Joanis, Kocmi, Koehn, Lo, Ljube{\v{s}}i{\'c}, Monz, Morishita, Nagata, Nakazawa, Pal, Post, and Zampieri}]{barrault-etal-2020-findings}
Lo{\"\i}c Barrault, Magdalena Biesialska, Ond{\v{r}}ej Bojar, Marta~R. Costa-juss{\`a}, Christian Federmann, Yvette Graham, Roman Grundkiewicz, Barry Haddow, Matthias Huck, Eric Joanis, Tom Kocmi, Philipp Koehn, Chi-kiu Lo, Nikola Ljube{\v{s}}i{\'c}, Christof Monz, Makoto Morishita, Masaaki Nagata, Toshiaki Nakazawa, Santanu Pal, Matt Post, and Marcos Zampieri. 2020.
\newblock \href {https://aclanthology.org/2020.wmt-1.1} {Findings of the 2020 conference on machine translation ({WMT}20)}.
\newblock In \emph{Proceedings of the Fifth Conference on Machine Translation}.

\bibitem[{Barrault et~al.(2019)Barrault, Bojar, Costa-juss{\`a}, Federmann, Fishel, Graham, Haddow, Huck, Koehn, Malmasi, Monz, M{\"u}ller, Pal, Post, and Zampieri}]{barrault-etal-2019-findings}
Lo{\"\i}c Barrault, Ond{\v{r}}ej Bojar, Marta~R. Costa-juss{\`a}, Christian Federmann, Mark Fishel, Yvette Graham, Barry Haddow, Matthias Huck, Philipp Koehn, Shervin Malmasi, Christof Monz, Mathias M{\"u}ller, Santanu Pal, Matt Post, and Marcos Zampieri. 2019.
\newblock \href {https://doi.org/10.18653/v1/W19-5301} {Findings of the 2019 conference on machine translation ({WMT}19)}.
\newblock In \emph{Proceedings of the Fourth Conference on Machine Translation (Volume 2: Shared Task Papers, Day 1)}.

\bibitem[{Bojar et~al.(2017)Bojar, Chatterjee, Federmann, Graham, Haddow, Huang, Huck, Koehn, Liu, Logacheva, Monz, Negri, Post, Rubino, Specia, and Turchi}]{bojar-etal-2017-findings}
Ond{\v{r}}ej Bojar, Rajen Chatterjee, Christian Federmann, Yvette Graham, Barry Haddow, Shujian Huang, Matthias Huck, Philipp Koehn, Qun Liu, Varvara Logacheva, Christof Monz, Matteo Negri, Matt Post, Raphael Rubino, Lucia Specia, and Marco Turchi. 2017.
\newblock \href {https://doi.org/10.18653/v1/W17-4717} {Findings of the 2017 conference on machine translation ({WMT}17)}.
\newblock In \emph{Proceedings of the Second Conference on Machine Translation}.

\bibitem[{Bojar et~al.(2018)Bojar, Federmann, Fishel, Graham, Haddow, Huck, Koehn, and Monz}]{bojar-etal-2018-findings}
Ond{\v{r}}ej Bojar, Christian Federmann, Mark Fishel, Yvette Graham, Barry Haddow, Matthias Huck, Philipp Koehn, and Christof Monz. 2018.
\newblock \href {https://doi.org/10.18653/v1/W18-6401} {Findings of the 2018 conference on machine translation ({WMT}18)}.
\newblock In \emph{Proceedings of the Third Conference on Machine Translation: Shared Task Papers}.

\bibitem[{Briakou et~al.(2023)Briakou, Cherry, and Foster}]{briakou-etal-2023-searching}
Eleftheria Briakou, Colin Cherry, and George Foster. 2023.
\newblock \href {https://doi.org/10.18653/v1/2023.acl-long.524} {Searching for needles in a haystack: On the role of incidental bilingualism in {P}a{LM}{'}s translation capability}.
\newblock In \emph{Proceedings of the 61st Annual Meeting of the Association for Computational Linguistics (Volume 1: Long Papers)}.

\bibitem[{Brown et~al.(2020)Brown, Mann, Ryder, Subbiah, Kaplan, Dhariwal, Neelakantan, Shyam, Sastry, Askell et~al.}]{Brown2020language}
Tom Brown, Benjamin Mann, Nick Ryder, Melanie Subbiah, Jared~D Kaplan, Prafulla Dhariwal, Arvind Neelakantan, Pranav Shyam, Girish Sastry, Amanda Askell, et~al. 2020.
\newblock \href {https://arxiv.org/abs/2005.14165} {Language models are few-shot learners}.
\newblock In \emph{Advances in Neural Information Processing Systems}.

\bibitem[{Chen et~al.(2024{\natexlab{a}})Chen, Li, Yan, Wang, Gunaratna, Yadav, Tang, Srinivasan, Zhou, Huang et~al.}]{chen2024alpagasus}
Lichang Chen, Shiyang Li, Jun Yan, Hai Wang, Kalpa Gunaratna, Vikas Yadav, Zheng Tang, Vijay Srinivasan, Tianyi Zhou, Heng Huang, et~al. 2024{\natexlab{a}}.
\newblock \href {https://openreview.net/pdf?id=FdVXgSJhvz} {Alpagasus: Training a better {Alpaca} model with fewer data}.
\newblock In \emph{The Twelfth International Conference on Learning Representations}.

\bibitem[{Chen et~al.(2024{\natexlab{b}})Chen, Guo, Haddow, and Heafield}]{chen2023iterative}
Pinzhen Chen, Zhicheng Guo, Barry Haddow, and Kenneth Heafield. 2024{\natexlab{b}}.
\newblock \href {https://aclanthology.org/2024.eamt-1.17} {Iterative translation refinement with large language models}.
\newblock In \emph{Proceedings of the 25th Annual Conference of the European Association for Machine Translation (Volume 1)}.

\bibitem[{Chen et~al.(2024{\natexlab{c}})Chen, Ji, Bogoychev, Kutuzov, Haddow, and Heafield}]{chen2023monolingual}
Pinzhen Chen, Shaoxiong Ji, Nikolay Bogoychev, Andrey Kutuzov, Barry Haddow, and Kenneth Heafield. 2024{\natexlab{c}}.
\newblock \href {https://aclanthology.org/2024.findings-eacl.90} {Monolingual or multilingual instruction tuning: Which makes a better {Alpaca}}.
\newblock In \emph{Findings of the Association for Computational Linguistics: EACL 2024}.

\bibitem[{Chiang et~al.(2023)Chiang, Li, Lin, Sheng, Wu, Zhang, Zheng, Zhuang, Zhuang, Gonzalez et~al.}]{vicuna2023}
Wei-Lin Chiang, Zhuohan Li, Zi~Lin, Ying Sheng, Zhanghao Wu, Hao Zhang, Lianmin Zheng, Siyuan Zhuang, Yonghao Zhuang, Joseph~E Gonzalez, et~al. 2023.
\newblock \href {https://lmsys.org/blog/2023-03-30-vicuna/} {Vicuna: An open-source chatbot impressing {GPT-4} with 90\%* {ChatGPT} quality}.
\newblock lmsys.org.

\bibitem[{Chirkova and Nikoulina(2024{\natexlab{a}})}]{chirkova2024key}
Nadezhda Chirkova and Vassilina Nikoulina. 2024{\natexlab{a}}.
\newblock \href {https://doi.org/10.18653/v1/2024.naacl-long.401} {Key ingredients for effective zero-shot cross-lingual knowledge transfer in generative tasks}.
\newblock In \emph{Proceedings of the 2024 Conference of the North American Chapter of the Association for Computational Linguistics: Human Language Technologies (Volume 1: Long Papers)}.

\bibitem[{Chirkova and Nikoulina(2024{\natexlab{b}})}]{chirkova2024zeroshot}
Nadezhda Chirkova and Vassilina Nikoulina. 2024{\natexlab{b}}.
\newblock \href {https://aclanthology.org/2024.inlg-main.53} {Zero-shot cross-lingual transfer in instruction tuning of large language models}.
\newblock In \emph{Proceedings of the 17th International Natural Language Generation Conference}.

\bibitem[{Choenni et~al.(2023)Choenni, Garrette, and Shutova}]{choenni-etal-2023-languages}
Rochelle Choenni, Dan Garrette, and Ekaterina Shutova. 2023.
\newblock \href {https://aclanthology.org/2023.emnlp-main.818/} {How do languages influence each other? studying cross-lingual data sharing during {LM} fine-tuning}.
\newblock In \emph{Proceedings of the 2023 Conference on Empirical Methods in Natural Language Processing}.

\bibitem[{Chung et~al.(2024)Chung, Hou, Longpre, Zoph, Tay, Fedus, Li, Wang, Dehghani, Brahma et~al.}]{chung2022scaling}
Hyung~Won Chung, Le~Hou, Shayne Longpre, Barret Zoph, Yi~Tay, William Fedus, Yunxuan Li, Xuezhi Wang, Mostafa Dehghani, Siddhartha Brahma, et~al. 2024.
\newblock \href {http://jmlr.org/papers/v25/23-0870.html} {Scaling instruction-finetuned language models}.
\newblock \emph{Journal of Machine Learning Research}.

\bibitem[{Ghazvininejad et~al.(2023)Ghazvininejad, Gonen, and Zettlemoyer}]{ghazvininejad2023dictionarybased}
Marjan Ghazvininejad, Hila Gonen, and Luke Zettlemoyer. 2023.
\newblock \href {https://arxiv.org/abs/2302.07856} {Dictionary-based phrase-level prompting of large language models for machine translation}.
\newblock \emph{arXiv preprint}.

\bibitem[{Ghosh et~al.(2024)Ghosh, Evuru, Kumar, S, Aneja, Jin, Duraiswami, and Manocha}]{ghosh2024closer}
Sreyan Ghosh, Chandra Kiran~Reddy Evuru, Sonal Kumar, Ramaneswaran S, Deepali Aneja, Zeyu Jin, Ramani Duraiswami, and Dinesh Manocha. 2024.
\newblock \href {https://proceedings.mlr.press/v235/ghosh24a.html} {A closer look at the limitations of instruction tuning}.
\newblock In \emph{Proceedings of the 41st International Conference on Machine Learning}.

\bibitem[{He et~al.(2024)He, Liang, Jiao, Zhang, Yang, Wang, Tu, Shi, and Wang}]{he2023exploring}
Zhiwei He, Tian Liang, Wenxiang Jiao, Zhuosheng Zhang, Yujiu Yang, Rui Wang, Zhaopeng Tu, Shuming Shi, and Xing Wang. 2024.
\newblock \href {https://doi.org/10.1162/tacl_a_00642} {Exploring human-like translation strategy with large language models}.
\newblock \emph{Transactions of the Association for Computational Linguistics}.

\bibitem[{Iyer et~al.(2023)Iyer, Chen, and Birch}]{iyer-etal-2023-towards}
Vivek Iyer, Pinzhen Chen, and Alexandra Birch. 2023.
\newblock \href {https://aclanthology.org/2023.wmt-1.44/} {Towards effective disambiguation for machine translation with large language models}.
\newblock In \emph{Proceedings of the Eighth Conference on Machine Translation}.

\bibitem[{Jiang et~al.(2023)Jiang, Sablayrolles, Mensch, Bamford, Chaplot, Casas, Bressand, Lengyel, Lample, Saulnier et~al.}]{jiang2023mistral}
Albert~Q Jiang, Alexandre Sablayrolles, Arthur Mensch, Chris Bamford, Devendra~Singh Chaplot, Diego de~las Casas, Florian Bressand, Gianna Lengyel, Guillaume Lample, Lucile Saulnier, et~al. 2023.
\newblock \href {https://arxiv.org/abs/2310.06825} {Mistral {7B}}.
\newblock \emph{arXiv preprint}.

\bibitem[{Jiao et~al.(2023)Jiao, Huang, Wang, He, Liang, Wang, Shi, and Tu}]{jiao2023_parrot}
Wenxiang Jiao, Jen-tse Huang, Wenxuan Wang, Zhiwei He, Tian Liang, Xing Wang, Shuming Shi, and Zhaopeng Tu. 2023.
\newblock \href {https://doi.org/10.18653/v1/2023.findings-emnlp.1001} {{P}arro{T}: Translating during chat using large language models tuned with human translation and feedback}.
\newblock In \emph{Findings of the Association for Computational Linguistics: EMNLP 2023}.

\bibitem[{Kew et~al.(2023)Kew, Schottmann, and Sennrich}]{kew2023turning}
Tannon Kew, Florian Schottmann, and Rico Sennrich. 2023.
\newblock \href {https://arxiv.org/abs/2312.12683} {Turning english-centric {LLMs} into polyglots: How much multilinguality is needed?}
\newblock \emph{arXiv preprint}.

\bibitem[{Kocmi et~al.(2022)Kocmi, Bawden, Bojar, Dvorkovich, Federmann, Fishel, Gowda, Graham, Grundkiewicz, Haddow, Knowles, Koehn, Monz, Morishita, Nagata, Nakazawa, Nov{\'a}k, Popel, and Popovi{\'c}}]{kocmi-etal-2022-findings}
Tom Kocmi, Rachel Bawden, Ond{\v{r}}ej Bojar, Anton Dvorkovich, Christian Federmann, Mark Fishel, Thamme Gowda, Yvette Graham, Roman Grundkiewicz, Barry Haddow, Rebecca Knowles, Philipp Koehn, Christof Monz, Makoto Morishita, Masaaki Nagata, Toshiaki Nakazawa, Michal Nov{\'a}k, Martin Popel, and Maja Popovi{\'c}. 2022.
\newblock \href {https://aclanthology.org/2022.wmt-1.1} {Findings of the 2022 conference on machine translation ({WMT}22)}.
\newblock In \emph{Proceedings of the Seventh Conference on Machine Translation (WMT)}.

\bibitem[{Li et~al.(2023)Li, Koto, Wu, Aji, and Baldwin}]{li2023bactrianx}
Haonan Li, Fajri Koto, Minghao Wu, Alham~Fikri Aji, and Timothy Baldwin. 2023.
\newblock \href {https://arxiv.org/abs/2305.15011} {{Bactrian-X}: Multilingual replicable instruction-following models with low-rank adaptation}.
\newblock \emph{arXiv preprint}.

\bibitem[{Li et~al.(2024)Li, Zhou, Huang, Cheng, and Chen}]{li2023eliciting}
Jiahuan Li, Hao Zhou, Shujian Huang, Shanbo Cheng, and Jiajun Chen. 2024.
\newblock \href {https://doi.org/10.1162/tacl_a_00655} {Eliciting the translation ability of large language models via multilingual finetuning with translation instructions}.
\newblock \emph{Transactions of the Association for Computational Linguistics}.

\bibitem[{Lin et~al.(2024)Lin, Ravichander, Lu, Dziri, Sclar, Chandu, Bhagavatula, and Choi}]{lin2024urial}
Bill~Yuchen Lin, Abhilasha Ravichander, Ximing Lu, Nouha Dziri, Melanie Sclar, Khyathi Chandu, Chandra Bhagavatula, and Yejin Choi. 2024.
\newblock \href {https://openreview.net/forum?id=wxJ0eXwwda} {Urial: Aligning untuned {LLM}s with just the 'write' amount of in-context learning}.
\newblock In \emph{The Twelfth International Conference on Learning Representations}.

\bibitem[{Mao and Yu(2024)}]{mao2024tuning}
Zhuoyuan Mao and Yen Yu. 2024.
\newblock \href {https://doi.org/10.18653/v1/2024.loresmt-1.1} {Tuning {LLM}s with contrastive alignment instructions for machine translation in unseen, low-resource languages}.
\newblock In \emph{Proceedings of the Seventh Workshop on Technologies for Machine Translation of Low-Resource Languages (LoResMT 2024)}.

\bibitem[{Mishra et~al.(2022)Mishra, Khashabi, Baral, and Hajishirzi}]{mishra-etal-2022-cross}
Swaroop Mishra, Daniel Khashabi, Chitta Baral, and Hannaneh Hajishirzi. 2022.
\newblock \href {https://doi.org/10.18653/v1/2022.acl-long.244} {Cross-task generalization via natural language crowdsourcing instructions}.
\newblock In \emph{Proceedings of the 60th Annual Meeting of the Association for Computational Linguistics (Volume 1: Long Papers)}.

\bibitem[{Mitchell et~al.(2024)Mitchell, Rafailov, Sharma, Finn, and Manning}]{mitchell2023emulator}
Eric Mitchell, Rafael Rafailov, Archit Sharma, Chelsea Finn, and Christopher~D Manning. 2024.
\newblock \href {https://openreview.net/forum?id=Eo7kv0sllr} {An emulator for fine-tuning large language models using small language models}.
\newblock In \emph{The Twelfth International Conference on Learning Representations}.

\bibitem[{Moslem et~al.(2023)Moslem, Haque, Kelleher, and Way}]{moslem-etal-2023-adaptive}
Yasmin Moslem, Rejwanul Haque, John~D. Kelleher, and Andy Way. 2023.
\newblock \href {https://aclanthology.org/2023.eamt-1.22} {Adaptive machine translation with large language models}.
\newblock In \emph{Proceedings of the 24th Annual Conference of the European Association for Machine Translation}.

\bibitem[{Muennighoff et~al.(2023)Muennighoff, Wang, Sutawika, Roberts, Biderman, Le~Scao, Bari, Shen, Yong, Schoelkopf, Tang, Radev, Aji, Almubarak, Albanie, Alyafeai, Webson, Raff, and Raffel}]{muennighoff-etal-2023-crosslingual}
Niklas Muennighoff, Thomas Wang, Lintang Sutawika, Adam Roberts, Stella Biderman, Teven Le~Scao, M~Saiful Bari, Sheng Shen, Zheng~Xin Yong, Hailey Schoelkopf, Xiangru Tang, Dragomir Radev, Alham~Fikri Aji, Khalid Almubarak, Samuel Albanie, Zaid Alyafeai, Albert Webson, Edward Raff, and Colin Raffel. 2023.
\newblock \href {https://doi.org/10.18653/v1/2023.acl-long.891} {Crosslingual generalization through multitask finetuning}.
\newblock In \emph{Proceedings of the 61st Annual Meeting of the Association for Computational Linguistics (Volume 1: Long Papers)}.

\bibitem[{Ouyang et~al.(2022)Ouyang, Wu, Jiang, Almeida, Wainwright, Mishkin, Zhang, Agarwal, Slama, Ray et~al.}]{ouyang2022_instructgpt}
Long Ouyang, Jeffrey Wu, Xu~Jiang, Diogo Almeida, Carroll Wainwright, Pamela Mishkin, Chong Zhang, Sandhini Agarwal, Katarina Slama, Alex Ray, et~al. 2022.
\newblock \href {http://papers.nips.cc/paper\_files/paper/2022/hash/b1efde53be364a73914f58805a001731-Abstract-Conference.html} {Training language models to follow instructions with human feedback}.
\newblock In \emph{Advances in Neural Information Processing Systems 35}.

\bibitem[{Papineni et~al.(2002)Papineni, Roukos, Ward, and Zhu}]{papineni-etal-2002-bleu}
Kishore Papineni, Salim Roukos, Todd Ward, and Wei-Jing Zhu. 2002.
\newblock \href {https://doi.org/10.3115/1073083.1073135} {{B}leu: a method for automatic evaluation of machine translation}.
\newblock In \emph{Proceedings of the 40th Annual Meeting of the Association for Computational Linguistics}.

\bibitem[{Post(2018)}]{post-2018-call}
Matt Post. 2018.
\newblock \href {https://doi.org/10.18653/v1/W18-6319} {A call for clarity in reporting {BLEU} scores}.
\newblock In \emph{Proceedings of the Third Conference on Machine Translation: Research Papers}.

\bibitem[{Radford et~al.(2019)Radford, Wu, Child, Luan, Amodei, Sutskever et~al.}]{radford2019language}
Alec Radford, Jeffrey Wu, Rewon Child, David Luan, Dario Amodei, Ilya Sutskever, et~al. 2019.
\newblock \href {https://d4mucfpksywv.cloudfront.net/better-language-models/language_models_are_unsupervised_multitask_learners.pdf} {Language models are unsupervised multitask learners}.
\newblock OpenAI blog.

\bibitem[{Raunak et~al.(2023)Raunak, Sharaf, Awadallah, and Menezes}]{raunak-etal-2023-leveraging}
Vikas Raunak, Amr Sharaf, Hany~Hassan Awadallah, and Arul Menezes. 2023.
\newblock \href {https://aclanthology.org/2023.findings-emnlp.804/} {Leveraging {GPT}-4 for automatic translation post-editing}.
\newblock In \emph{Findings of the Association for Computational Linguistics: EMNLP 2023}.

\bibitem[{Rei et~al.(2020)Rei, Stewart, Farinha, and Lavie}]{rei-etal-2020-comet}
Ricardo Rei, Craig Stewart, Ana~C Farinha, and Alon Lavie. 2020.
\newblock \href {https://doi.org/10.18653/v1/2020.emnlp-main.213} {{COMET}: A neural framework for {MT} evaluation}.
\newblock In \emph{Proceedings of the 2020 Conference on Empirical Methods in Natural Language Processing (EMNLP)}.

\bibitem[{Sanh et~al.(2022)Sanh, Webson, Raffel, Bach, Sutawika, Alyafeai, Chaffin, Stiegler, Scao, Raja et~al.}]{sanh2022multitask}
Victor Sanh, Albert Webson, Colin Raffel, Stephen~H Bach, Lintang Sutawika, Zaid Alyafeai, Antoine Chaffin, Arnaud Stiegler, Teven~Le Scao, Arun Raja, et~al. 2022.
\newblock \href {https://arxiv.org/abs/2110.08207} {Multitask prompted training enables zero-shot task generalization}.
\newblock In \emph{International Conference on Learning Representations}.

\bibitem[{Sarti et~al.(2023)Sarti, Htut, Niu, Hsu, Currey, Dinu, and Nadejde}]{sarti-etal-2023-ramp}
Gabriele Sarti, Phu~Mon Htut, Xing Niu, Benjamin Hsu, Anna Currey, Georgiana Dinu, and Maria Nadejde. 2023.
\newblock \href {https://doi.org/10.18653/v1/2023.acl-short.126} {{RAMP}: Retrieval and attribute-marking enhanced prompting for attribute-controlled translation}.
\newblock In \emph{Proceedings of the 61st Annual Meeting of the Association for Computational Linguistics (Volume 2: Short Papers)}.

\bibitem[{Sennrich et~al.(2016)Sennrich, Haddow, and Birch}]{sennrich-etal-2016-improving}
Rico Sennrich, Barry Haddow, and Alexandra Birch. 2016.
\newblock \href {https://doi.org/10.18653/v1/P16-1009} {Improving neural machine translation models with monolingual data}.
\newblock In \emph{Proceedings of the 54th Annual Meeting of the Association for Computational Linguistics (Volume 1: Long Papers)}.

\bibitem[{Shaham et~al.(2024)Shaham, Herzig, Aharoni, Szpektor, Tsarfaty, and Eyal}]{shaham2024multilingual}
Uri Shaham, Jonathan Herzig, Roee Aharoni, Idan Szpektor, Reut Tsarfaty, and Matan Eyal. 2024.
\newblock \href {https://doi.org/10.18653/v1/2024.findings-acl.136} {Multilingual instruction tuning with just a pinch of multilinguality}.
\newblock In \emph{Findings of the Association for Computational Linguistics ACL 2024}.

\bibitem[{Sia et~al.(2024)Sia, Mueller, and Duh}]{Sia2024WhereDI}
Suzanna Sia, David Mueller, and Kevin Duh. 2024.
\newblock \href {https://arxiv.org/abs/2403.04510} {Where does in-context translation happen in large language models}.
\newblock \emph{arXiv preprint}.

\bibitem[{Stap et~al.(2024)Stap, Hasler, Byrne, Monz, and Tran}]{Stap2024}
David Stap, Eva Hasler, Bill Byrne, Christof Monz, and Ke~Tran. 2024.
\newblock \href {https://doi.org/10.18653/v1/2024.acl-long.336} {The fine-tuning paradox: Boosting translation quality without sacrificing {LLM} abilities}.
\newblock In \emph{Proceedings of the 62nd Annual Meeting of the Association for Computational Linguistics (Volume 1: Long Papers)}.

\bibitem[{Taori et~al.(2023)Taori, Gulrajani, Zhang, Dubois, Li, Guestrin, Liang, and Hashimoto}]{Taori2023}
Rohan Taori, Ishaan Gulrajani, Tianyi Zhang, Yann Dubois, Xuechen Li, Carlos Guestrin, Percy Liang, and Tatsunori~B. Hashimoto. 2023.
\newblock \href {https://github.com/tatsu-lab/stanford_alpaca} {{Stanford} {Alpaca}: An instruction-following {LLaMA} model}.
\newblock GitHub repository.

\bibitem[{Tiedemann and Thottingal(2020)}]{tiedemann-thottingal-2020-opus}
J{\"o}rg Tiedemann and Santhosh Thottingal. 2020.
\newblock \href {https://aclanthology.org/2020.eamt-1.61} {{OPUS}-{MT} {--} building open translation services for the world}.
\newblock In \emph{Proceedings of the 22nd Annual Conference of the European Association for Machine Translation}.

\bibitem[{Touvron et~al.(2023)Touvron, Martin, Stone, Albert, Almahairi, Babaei, Bashlykov, Batra, Bhargava, Bhosale et~al.}]{touvron2023llama2}
Hugo Touvron, Louis Martin, Kevin Stone, Peter Albert, Amjad Almahairi, Yasmine Babaei, Nikolay Bashlykov, Soumya Batra, Prajjwal Bhargava, Shruti Bhosale, et~al. 2023.
\newblock \href {https://arxiv.org/abs/2307.09288} {Llama 2: Open foundation and fine-tuned chat models}.
\newblock \emph{arXiv preprint}.

\bibitem[{{\"U}st{\"u}n et~al.(2024){\"U}st{\"u}n, Aryabumi, Yong, Ko, D'souza, Onilude, Bhandari, Singh, Ooi, Kayid et~al.}]{ustun2024aya}
Ahmet {\"U}st{\"u}n, Viraat Aryabumi, Zheng-Xin Yong, Wei-Yin Ko, Daniel D'souza, Gbemileke Onilude, Neel Bhandari, Shivalika Singh, Hui-Lee Ooi, Amr Kayid, et~al. 2024.
\newblock \href {https://doi.org/10.18653/v1/2024.acl-long.845} {Aya model: An instruction finetuned open-access multilingual language model}.
\newblock In \emph{Proceedings of the 62nd Annual Meeting of the Association for Computational Linguistics (Volume 1: Long Papers)}.

\bibitem[{Vilar et~al.(2023)Vilar, Freitag, Cherry, Luo, Ratnakar, and Foster}]{vilar-etal-2023-prompting}
David Vilar, Markus Freitag, Colin Cherry, Jiaming Luo, Viresh Ratnakar, and George Foster. 2023.
\newblock \href {https://doi.org/10.18653/v1/2023.acl-long.859} {Prompting {P}a{LM} for translation: Assessing strategies and performance}.
\newblock In \emph{Proceedings of the 61st Annual Meeting of the Association for Computational Linguistics (Volume 1: Long Papers)}.

\bibitem[{Wei et~al.(2022)Wei, Bosma, Zhao, Guu, Yu, Lester, Du, Dai, and Le}]{wei2022finetuned}
Jason Wei, Maarten Bosma, Vincent Zhao, Kelvin Guu, Adams~Wei Yu, Brian Lester, Nan Du, Andrew~M. Dai, and Quoc~V Le. 2022.
\newblock \href {https://arxiv.org/abs/2109.01652} {Finetuned language models are zero-shot learners}.
\newblock In \emph{International Conference on Learning Representations}.

\bibitem[{Wu et~al.(2024{\natexlab{a}})Wu, Tan, Meng, Stap, and Monz}]{wu-etal-2024-far}
Di~Wu, Shaomu Tan, Yan Meng, David Stap, and Christof Monz. 2024{\natexlab{a}}.
\newblock \href {https://aclanthology.org/2024.findings-acl.896} {How far can 100 samples go? unlocking zero-shot translation with tiny multi-parallel data}.
\newblock In \emph{Findings of the Association for Computational Linguistics ACL 2024}.

\bibitem[{Wu et~al.(2024{\natexlab{b}})Wu, Vu, Qu, Foster, and Haffari}]{wu2024adapting}
Minghao Wu, Thuy-Trang Vu, Lizhen Qu, George Foster, and Gholamreza Haffari. 2024{\natexlab{b}}.
\newblock \href {https://arxiv.org/abs/2401.06468} {Adapting large language models for document-level machine translation}.
\newblock \emph{arXiv preprint}.

\bibitem[{Wu et~al.(2024{\natexlab{c}})Wu, Waheed, Zhang, Abdul-Mageed, and Aji}]{wu2024laminilm}
Minghao Wu, Abdul Waheed, Chiyu Zhang, Muhammad Abdul-Mageed, and Alham Aji. 2024{\natexlab{c}}.
\newblock \href {https://aclanthology.org/2024.eacl-long.57} {{L}a{M}ini-{LM}: A diverse herd of distilled models from large-scale instructions}.
\newblock In \emph{Proceedings of the 18th Conference of the European Chapter of the Association for Computational Linguistics}.

\bibitem[{Wu and Dredze(2019)}]{wu-dredze-2019-beto}
Shijie Wu and Mark Dredze. 2019.
\newblock \href {https://doi.org/10.18653/v1/D19-1077} {Beto, bentz, becas: The surprising cross-lingual effectiveness of {BERT}}.
\newblock In \emph{Proceedings of the 2019 Conference on Empirical Methods in Natural Language Processing and the 9th International Joint Conference on Natural Language Processing (EMNLP-IJCNLP)}.

\bibitem[{Xu et~al.(2024{\natexlab{a}})Xu, Kim, Sharaf, and Awadalla}]{xu2024alma}
Haoran Xu, Young~Jin Kim, Amr Sharaf, and Hany~Hassan Awadalla. 2024{\natexlab{a}}.
\newblock \href {https://openreview.net/forum?id=farT6XXntP} {A paradigm shift in machine translation: Boosting translation performance of large language models}.
\newblock In \emph{The Twelfth International Conference on Learning Representations}.

\bibitem[{Xu et~al.(2024{\natexlab{b}})Xu, Sharaf, Chen, Tan, Shen, Van~Durme, Murray, and Kim}]{xu2024contrastive}
Haoran Xu, Amr Sharaf, Yunmo Chen, Weiting Tan, Lingfeng Shen, Benjamin Van~Durme, Kenton Murray, and Young~Jin Kim. 2024{\natexlab{b}}.
\newblock \href {https://proceedings.mlr.press/v235/xu24t.html} {Contrastive preference optimization: Pushing the boundaries of {LLM} performance in machine translation}.
\newblock In \emph{Proceedings of the 41st International Conference on Machine Learning}.

\bibitem[{Zeng et~al.(2024)Zeng, Meng, Yin, and Zhou}]{zeng2024_tim}
Jiali Zeng, Fandong Meng, Yongjing Yin, and Jie Zhou. 2024.
\newblock \href {https://ojs.aaai.org/index.php/AAAI/article/view/29920} {Teaching large language models to translate with comparison}.
\newblock In \emph{Proceedings of the AAAI Conference on Artificial Intelligence}.

\bibitem[{Zhang et~al.(2023{\natexlab{a}})Zhang, Haddow, and Birch}]{zhang2023prompting}
Biao Zhang, Barry Haddow, and Alexandra Birch. 2023{\natexlab{a}}.
\newblock \href {https://arxiv.org/abs/2301.07069} {Prompting large language model for machine translation: a case study}.
\newblock In \emph{Proceedings of the 40th International Conference on Machine Learning}.

\bibitem[{Zhang et~al.(2024)Zhang, Liu, Cherry, and Firat}]{zhang2024when}
Biao Zhang, Zhongtao Liu, Colin Cherry, and Orhan Firat. 2024.
\newblock \href {https://arxiv.org/abs/2402.17193} {When scaling meets {LLM} finetuning: The effect of data, model and finetuning method}.
\newblock In \emph{The Twelfth International Conference on Learning Representations}.

\bibitem[{Zhang et~al.(2023{\natexlab{b}})Zhang, Fang, Zhang, Ma, Zhou, Huang, Bu, Gui, Chen, Chen et~al.}]{zhang2023bayling}
Shaolei Zhang, Qingkai Fang, Zhuocheng Zhang, Zhengrui Ma, Yan Zhou, Langlin Huang, Mengyu Bu, Shangtong Gui, Yunji Chen, Xilin Chen, et~al. 2023{\natexlab{b}}.
\newblock \href {https://arxiv.org/abs/2306.10968} {{BayLing}: Bridging cross-lingual alignment and instruction following through interactive translation for large language models}.
\newblock \emph{arXiv preprint}.

\bibitem[{Zhang et~al.(2023{\natexlab{c}})Zhang, Rajabi, Duh, and Koehn}]{zhang-etal-2023-machine}
Xuan Zhang, Navid Rajabi, Kevin Duh, and Philipp Koehn. 2023{\natexlab{c}}.
\newblock \href {https://doi.org/10.18653/v1/2023.wmt-1.43} {Machine translation with large language models: Prompting, few-shot learning, and fine-tuning with {QL}o{RA}}.
\newblock In \emph{Proceedings of the Eighth Conference on Machine Translation}.

\bibitem[{Zhou et~al.(2023)Zhou, Liu, Xu, Iyer, Sun, Mao, Ma, Efrat, Yu, Yu et~al.}]{zhou2023lima}
Chunting Zhou, Pengfei Liu, Puxin Xu, Srinivasan Iyer, Jiao Sun, Yuning Mao, Xuezhe Ma, Avia Efrat, Ping Yu, Lili Yu, et~al. 2023.
\newblock \href {https://arxiv.org/abs/2305.11206} {{LIMA}: Less is more for alignment}.
\newblock In \emph{Thirty-seventh Conference on Neural Information Processing Systems}.

\bibitem[{Zhu et~al.(2024)Zhu, Trenous, Shen, Klakow, Byrne, and Hasler}]{zhu-preference-mt}
Dawei Zhu, Sony Trenous, Xiaoyu Shen, Dietrich Klakow, Bill Byrne, and Eva Hasler. 2024.
\newblock \href {https://doi.org/10.18653/v1/2024.naacl-long.186} {A preference-driven paradigm for enhanced translation with large language models}.
\newblock In \emph{Proceedings of the 2024 Conference of the North American Chapter of the Association for Computational Linguistics: Human Language Technologies}.

\end{thebibliography}

\newpage
\clearpage
\appendix

\FloatBarrier
\section{Model Performance with Varying Training Sample Sizes}
\label{appendix:sec:model_performance_sample_size_sep}
\begin{figure*}[ht]
    \centering
    \includegraphics[width=0.40\textwidth]{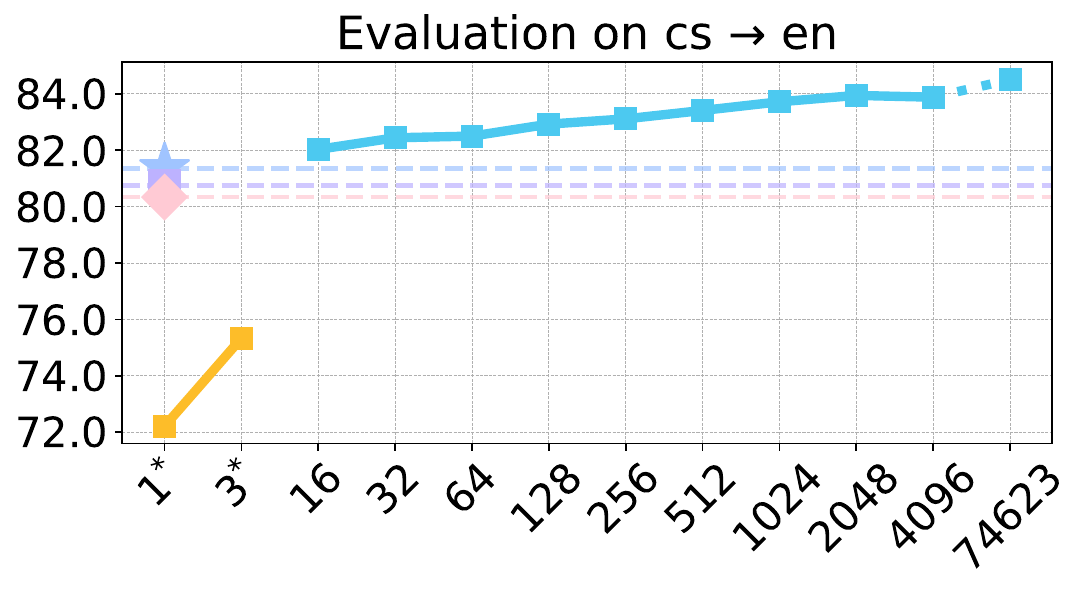}
    \includegraphics[width=0.40\textwidth]{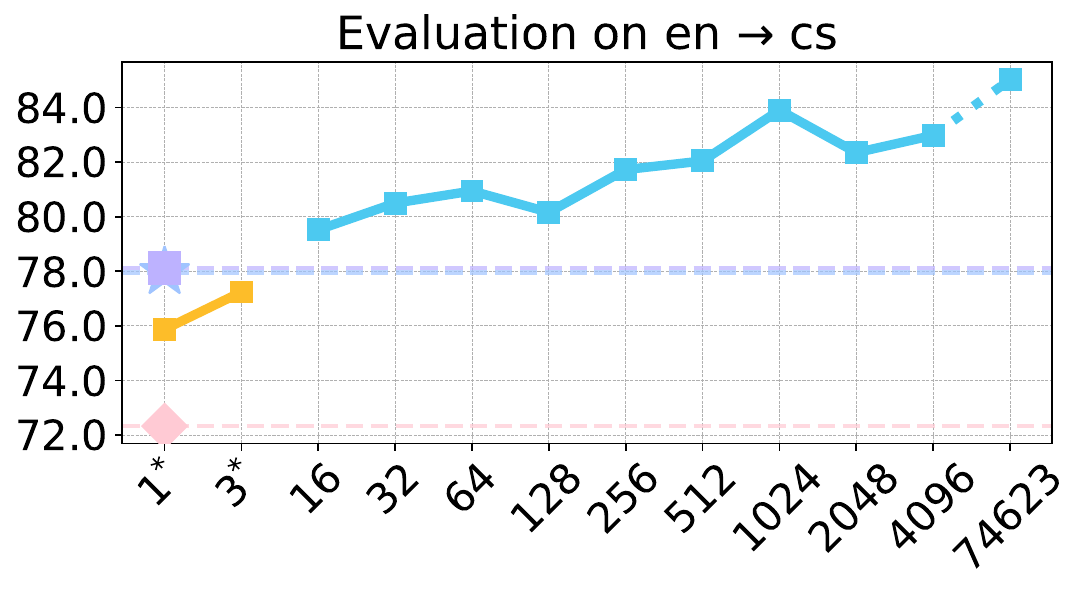}
    \includegraphics[width=0.40\textwidth]{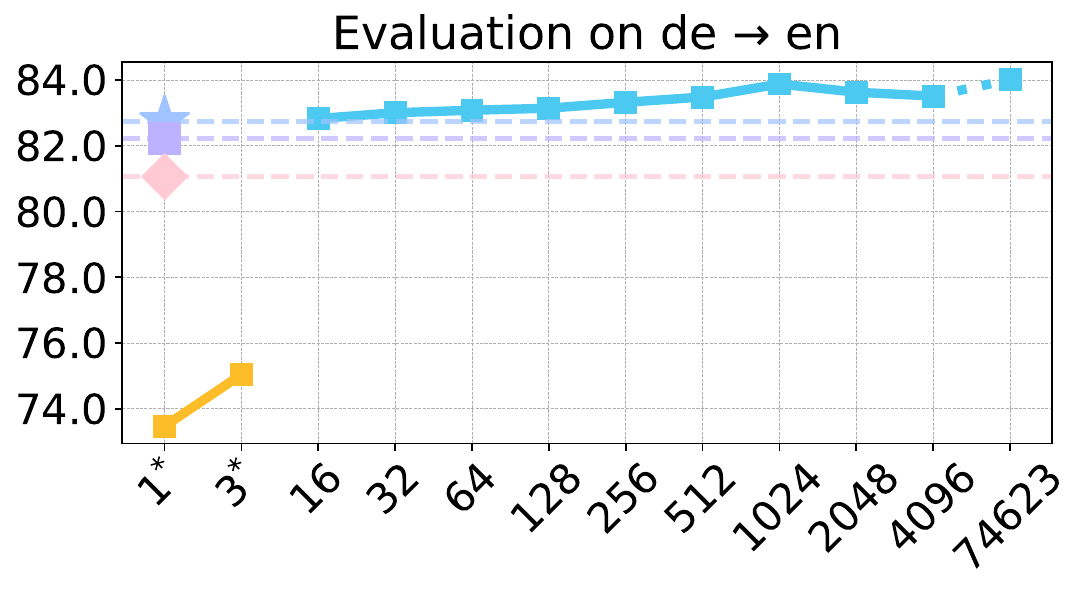}
    \includegraphics[width=0.40\textwidth]{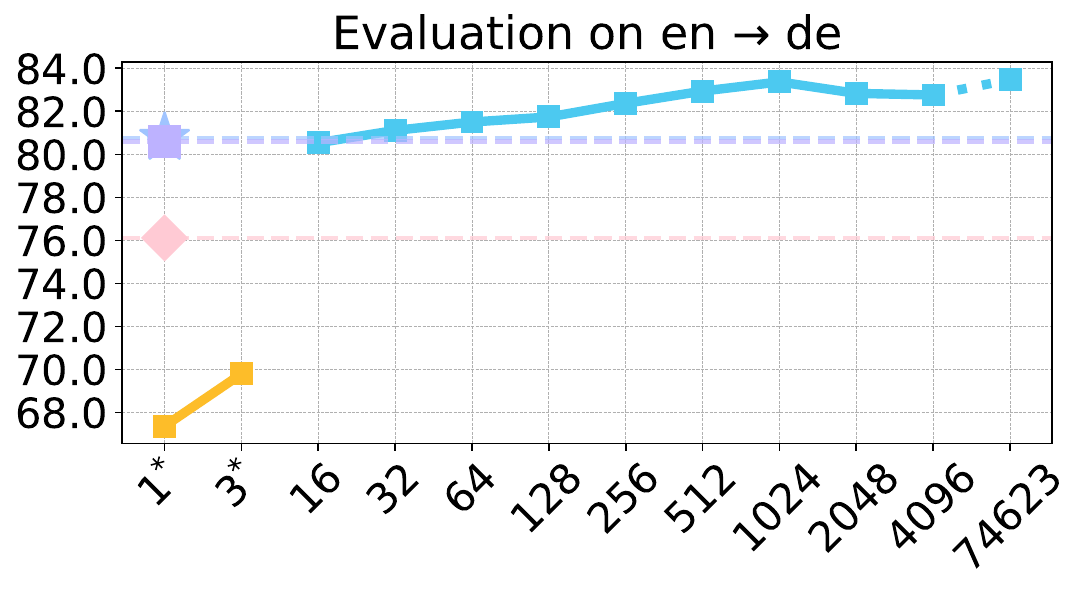}
    \includegraphics[width=0.40\textwidth]{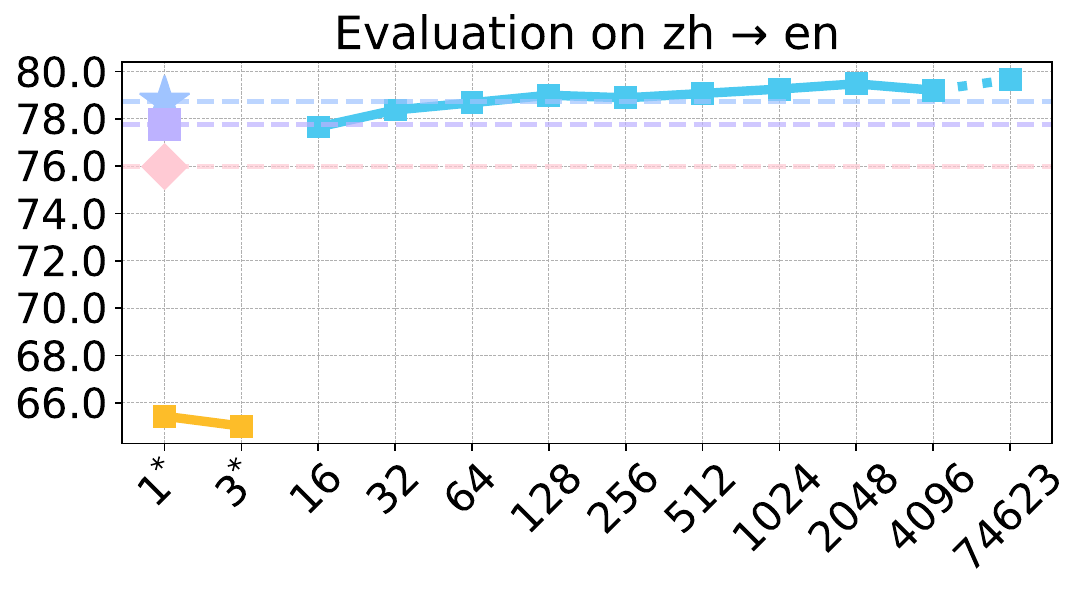}
    \includegraphics[width=0.40\textwidth]{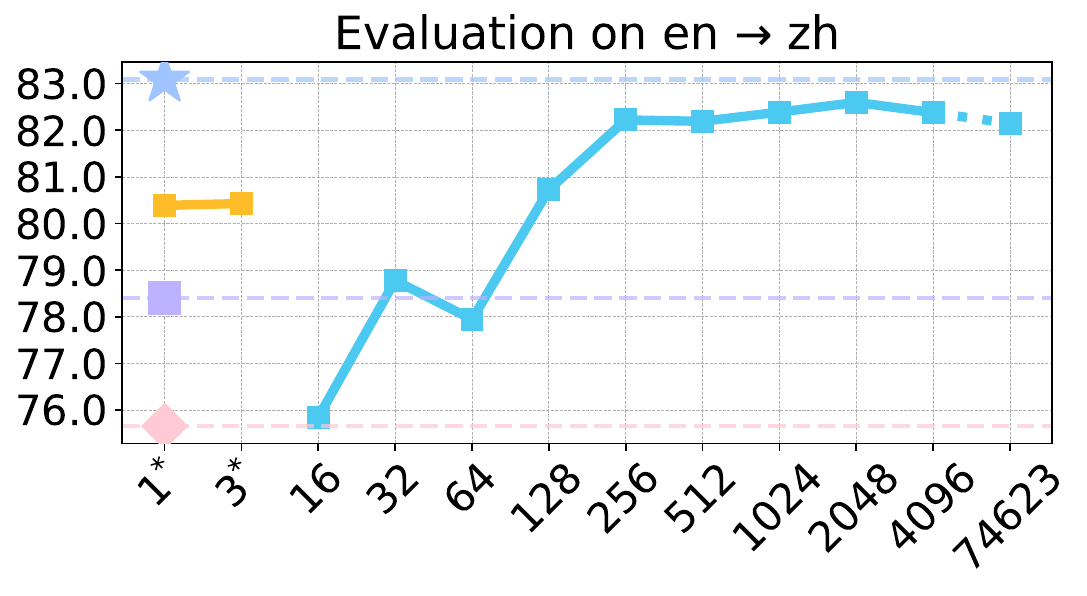}
    \includegraphics[width=0.40\textwidth]{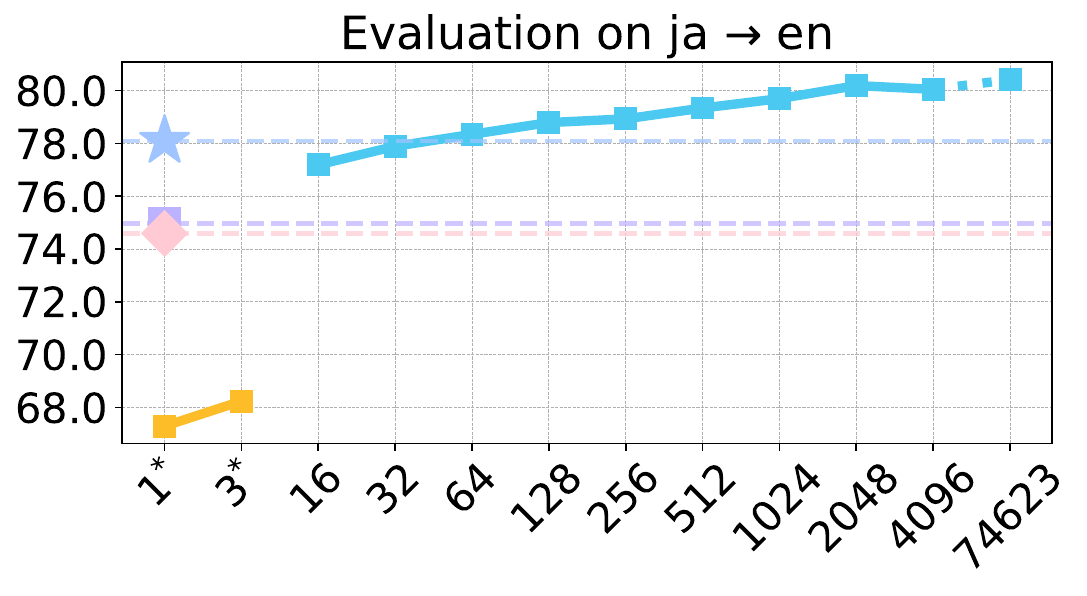}
    \includegraphics[width=0.40\textwidth]{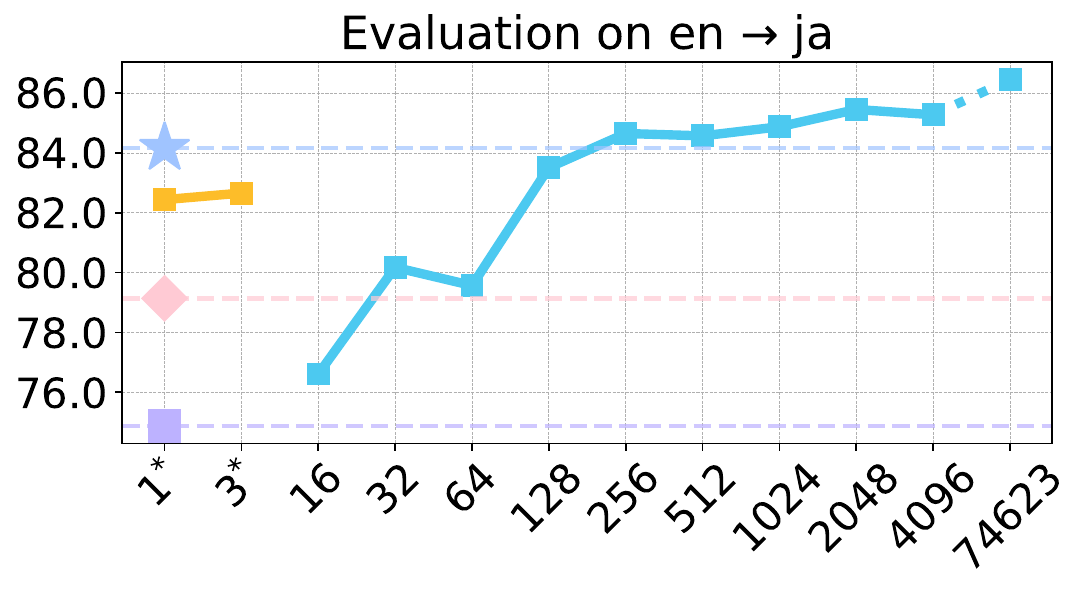}
    \includegraphics[width=0.40\textwidth]{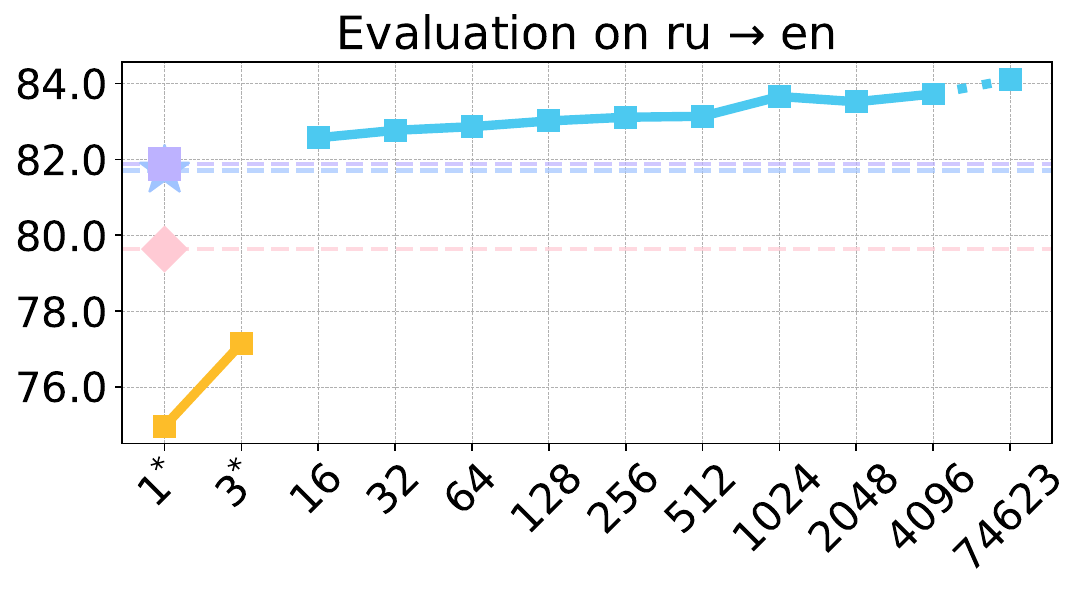}
    \includegraphics[width=0.40\textwidth]{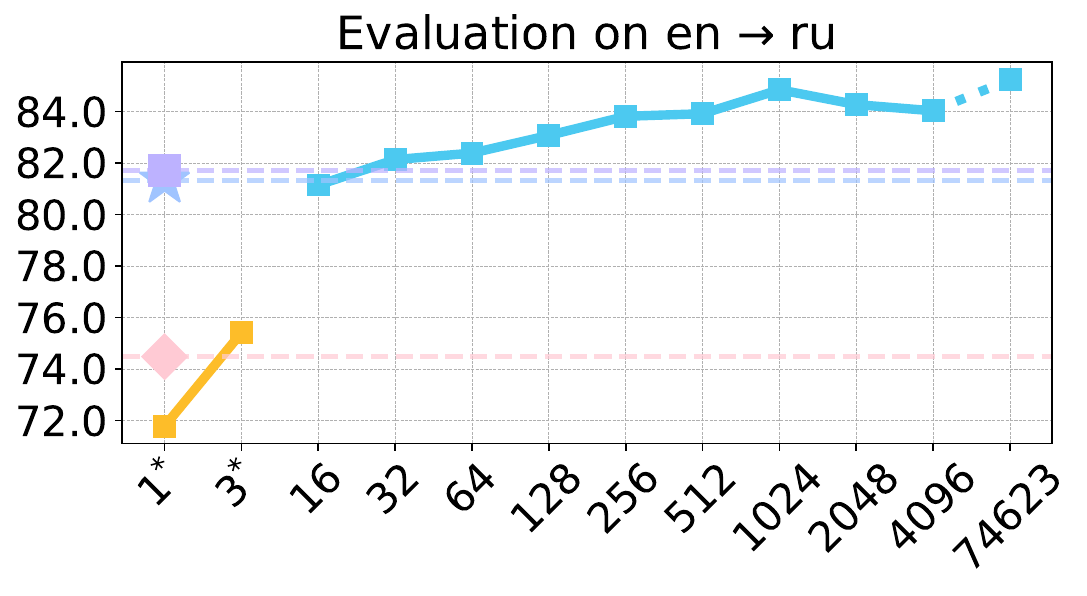}
    \includegraphics[width=0.63\textwidth]{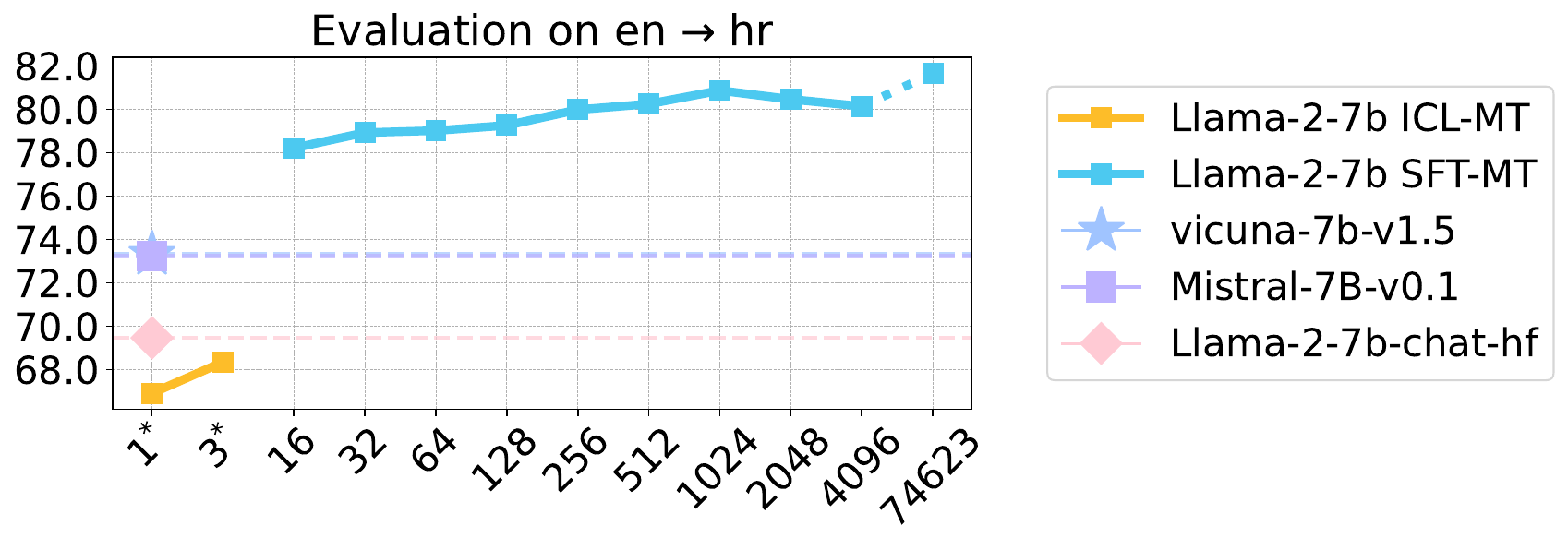}
    \caption{COMET scores between instruction-tuned baselines and our models at different training data sizes, evaluated on individual translation directions. ICL is used for training sizes at or below 3, indicated with "$^*$"; otherwise, we perform SFT. With only 32 examples for SFT, Llama-2 outperforms general-purpose, instruction-tuned baselines. Base.: instruction-tuned baseline models.}
    \label{fig:size_exp_comet_sep}
\end{figure*}

\begin{figure*}[ht]
    \centering
    \includegraphics[width=0.40\textwidth]{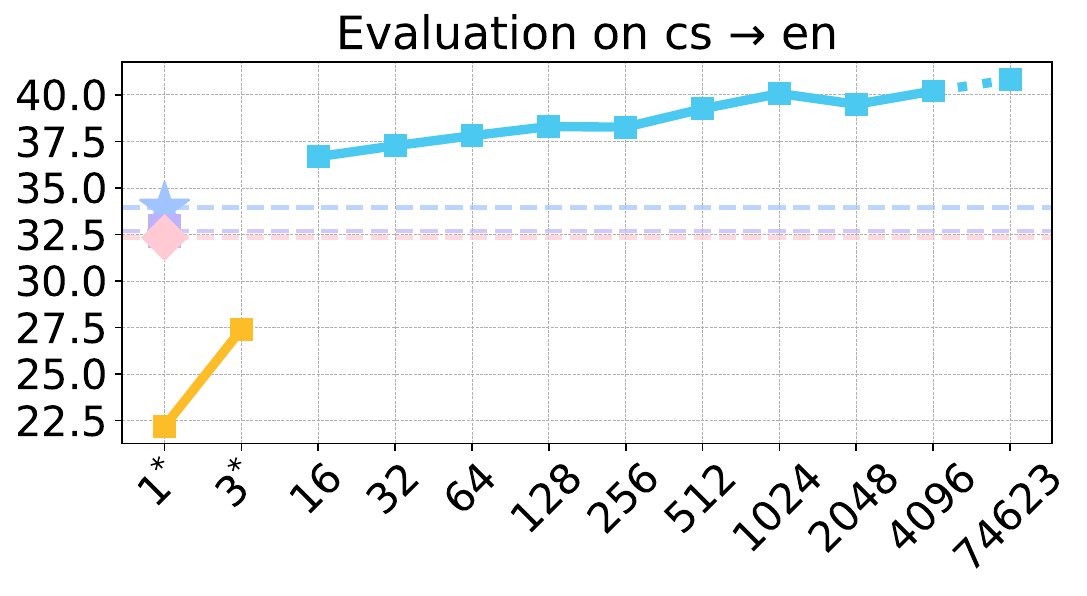}
    \includegraphics[width=0.40\textwidth]{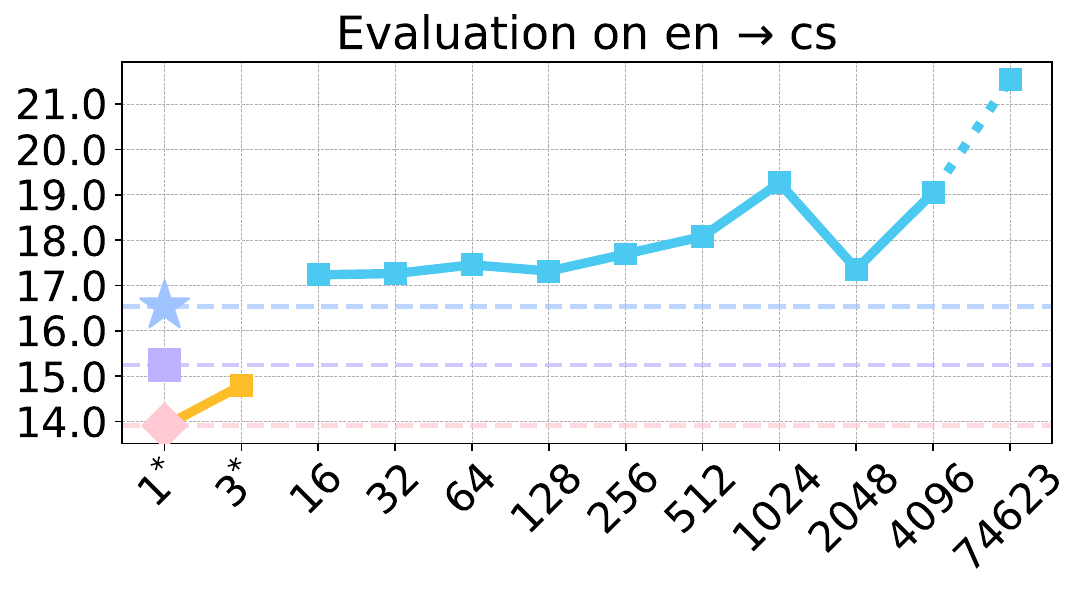}
    \includegraphics[width=0.40\textwidth]{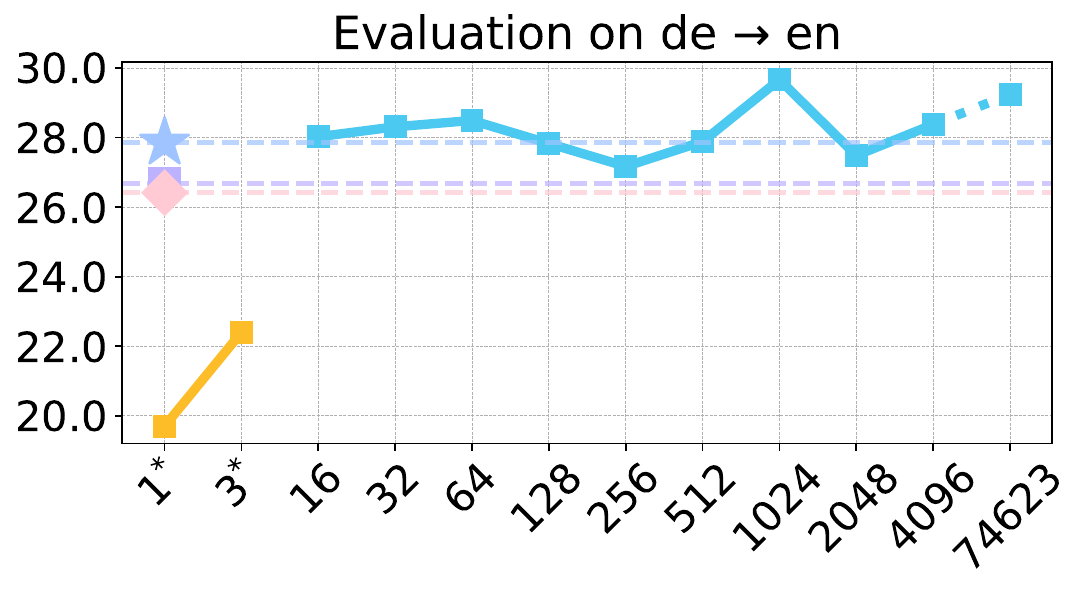}
    \includegraphics[width=0.40\textwidth]{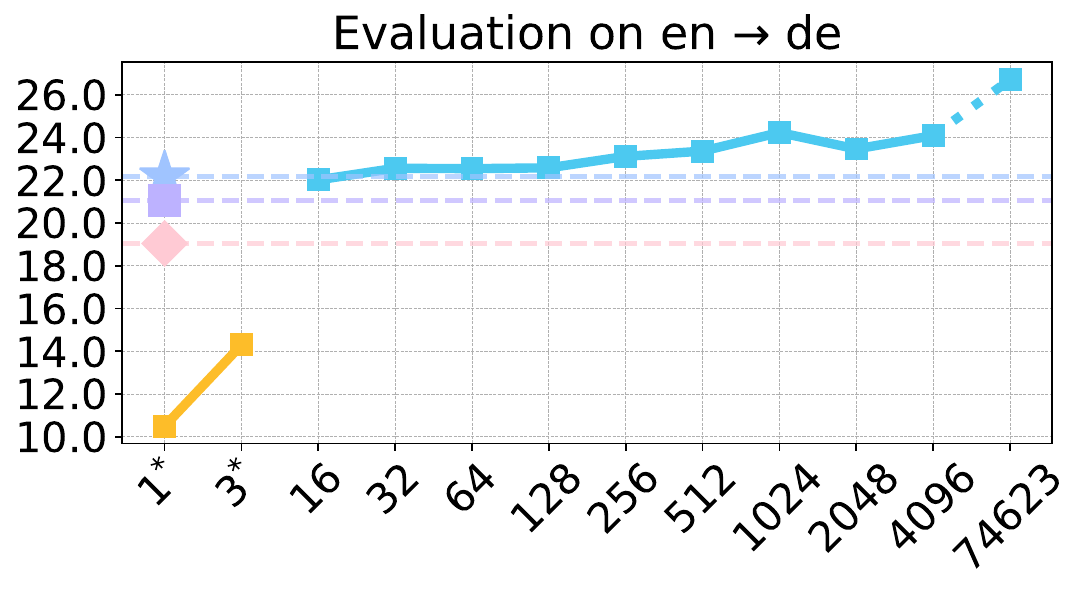}
    \includegraphics[width=0.40\textwidth]{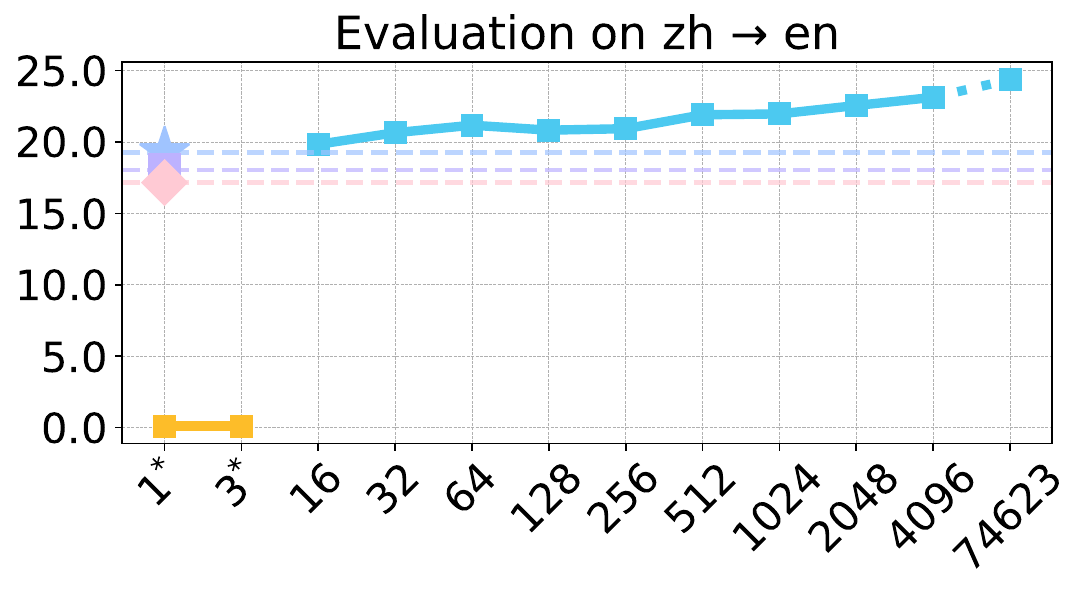}
    \includegraphics[width=0.40\textwidth]{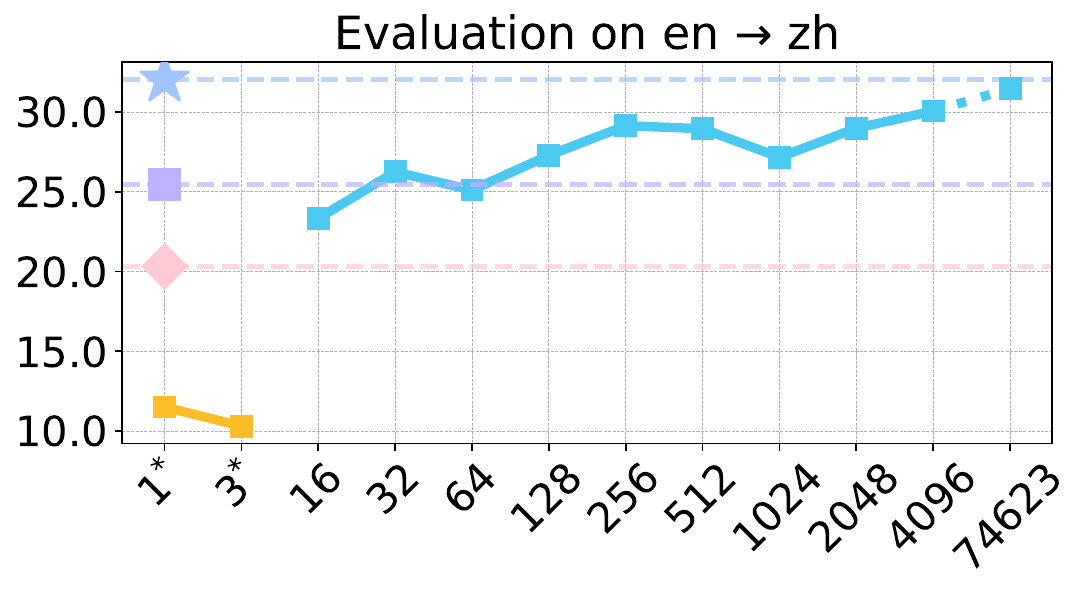}
    \includegraphics[width=0.40\textwidth]{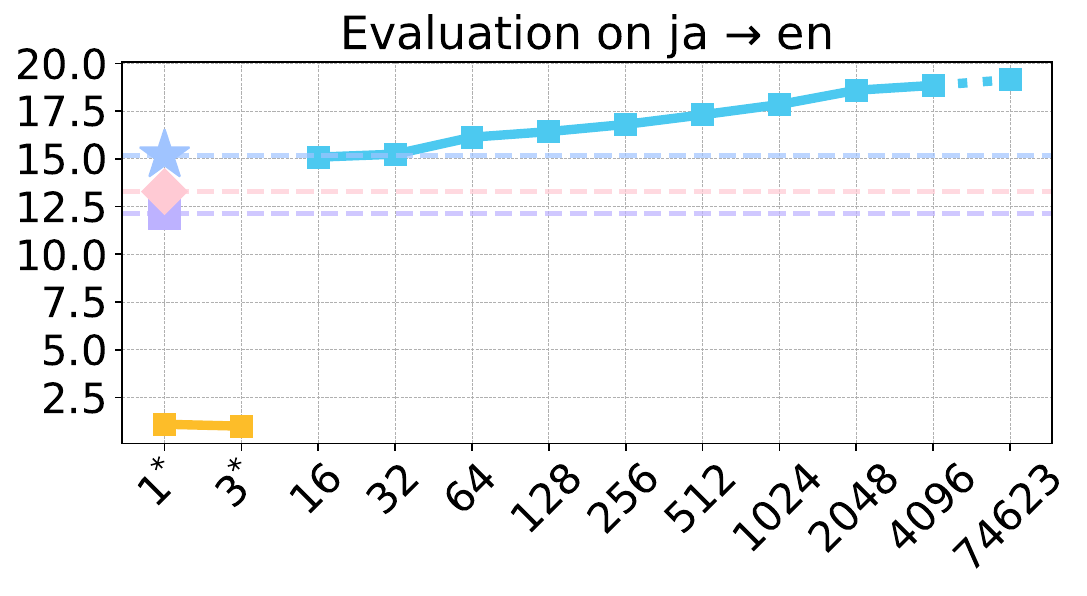}
    \includegraphics[width=0.40\textwidth]{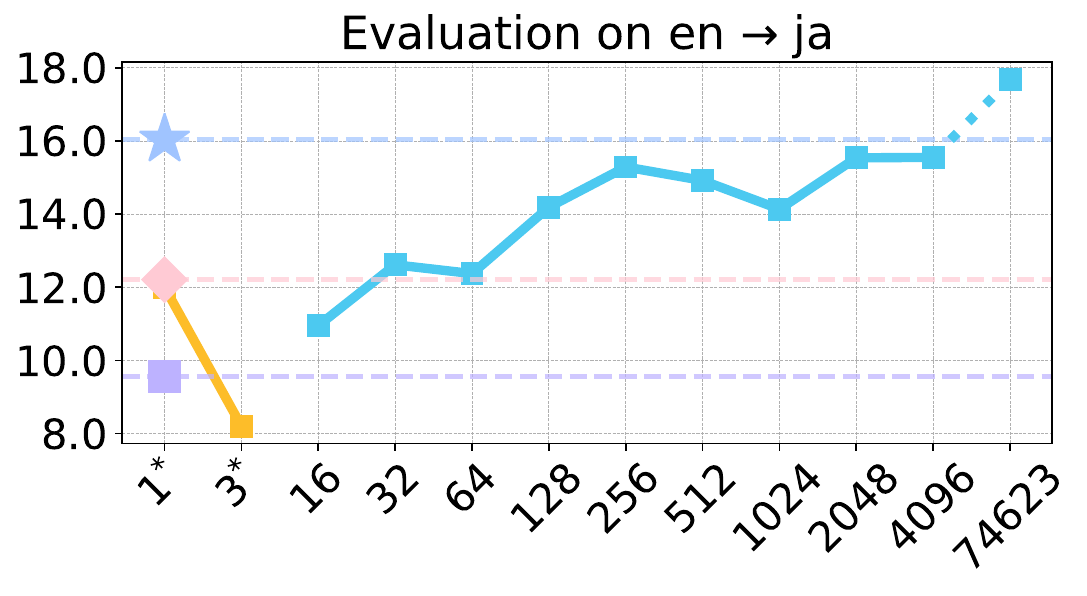}
    \includegraphics[width=0.40\textwidth]{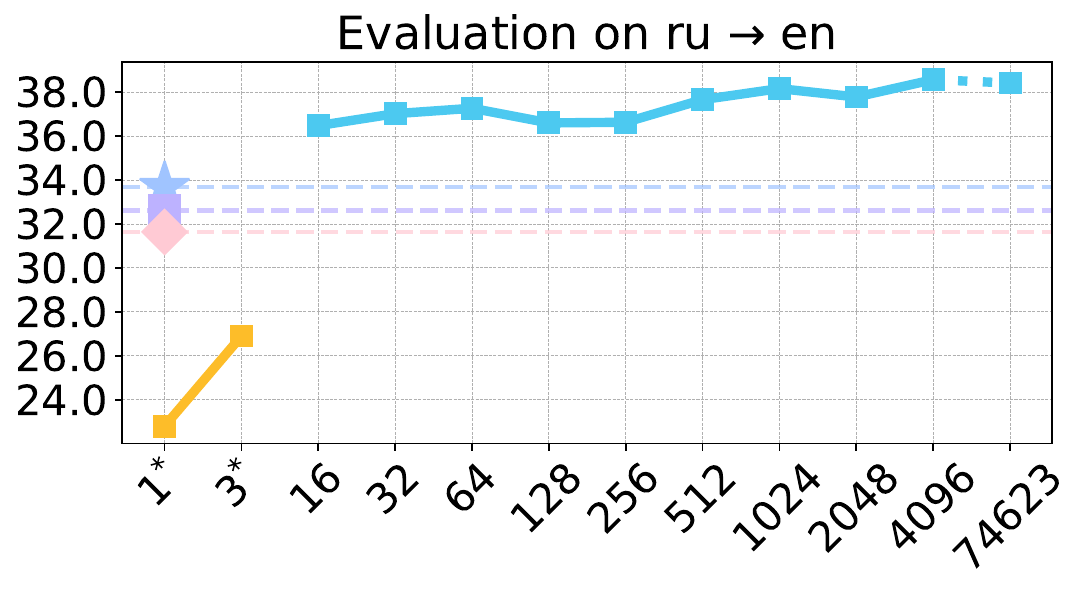}
    \includegraphics[width=0.40\textwidth]{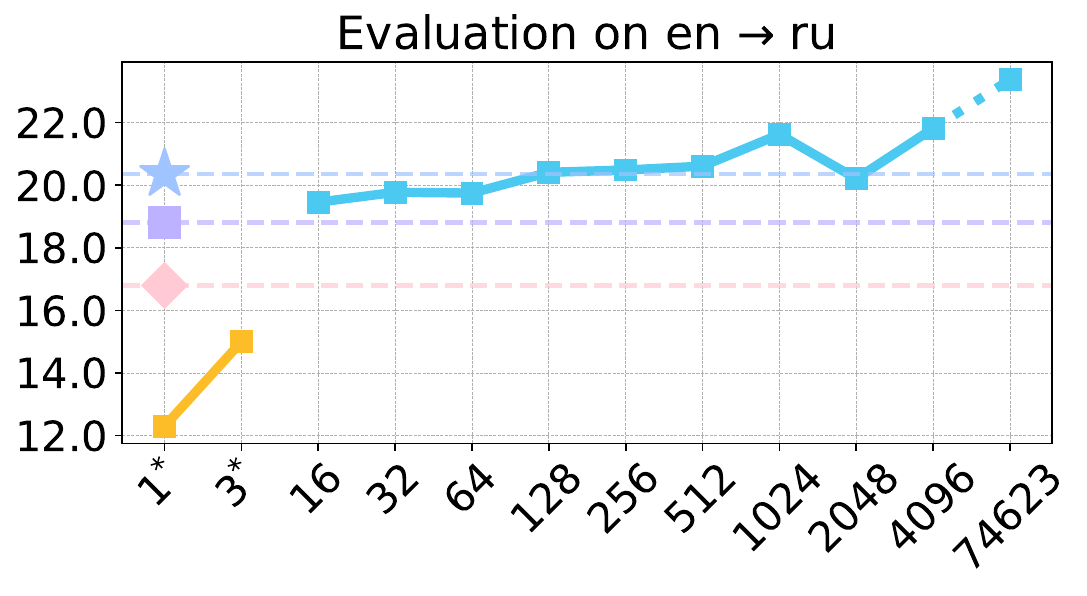}
    \includegraphics[width=0.63\textwidth]{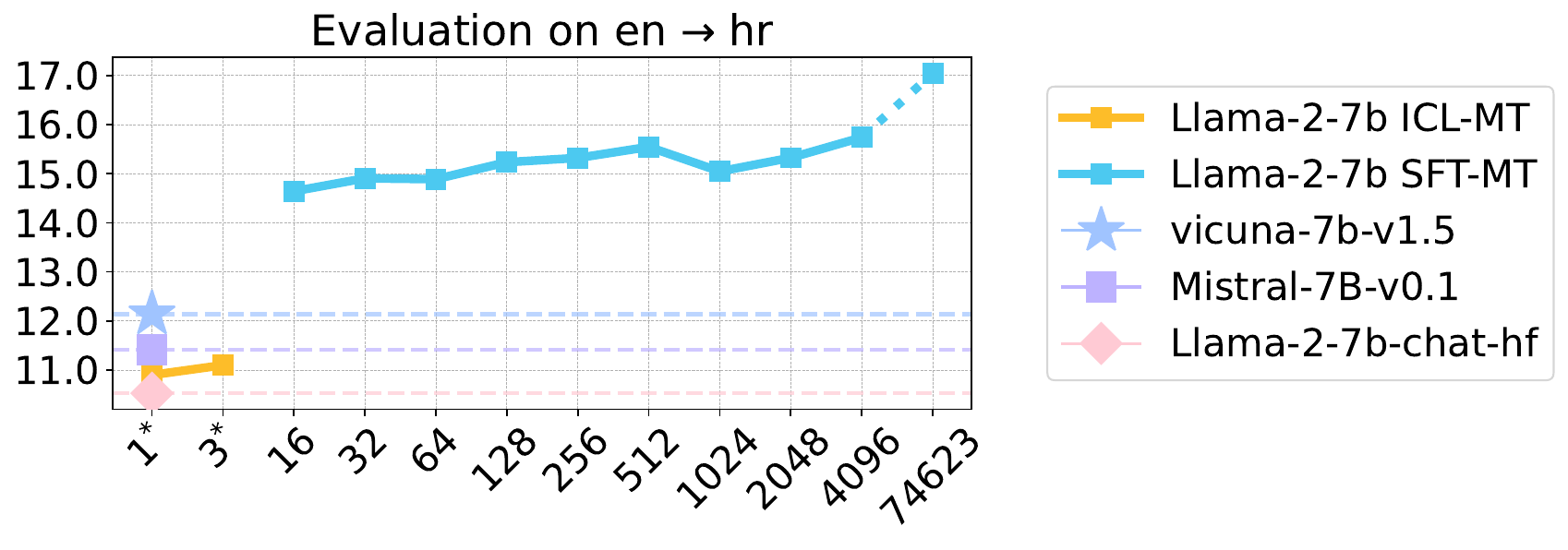}
    \caption{BLEU scores between instruction-tuned baselines and our models at different training data sizes, evaluated on individual translation directions. ICL is used for training sizes at or below 3, indicated with "$^*$"; otherwise, we perform SFT. With only 32 examples for SFT, Llama-2 outperforms general-purpose, instruction-tuned baselines. Base.: instruction-tuned baseline models.}
    \label{fig:size_exp_bleu_sep}
\end{figure*}

In Figure~\ref{fig:size_exp_comet_sep} and Figure~\ref{fig:size_exp_bleu_sep}, we present the performance for instruction-tuned baselines and our models on different evaluation directions. For most directions, using only $32$ training samples can achieve competitive performance and beat all three instruction-tuned baselines. There are several exceptional cases, including en$\rightarrow$zh and en$\rightarrow$ja, in which the COMET score of SFT with a limited number of samples ($32$ or $64$) is worse than 1-shot in-context learning.

\begin{figure*}[ht]
    \centering
    \includegraphics[width=0.8\linewidth]{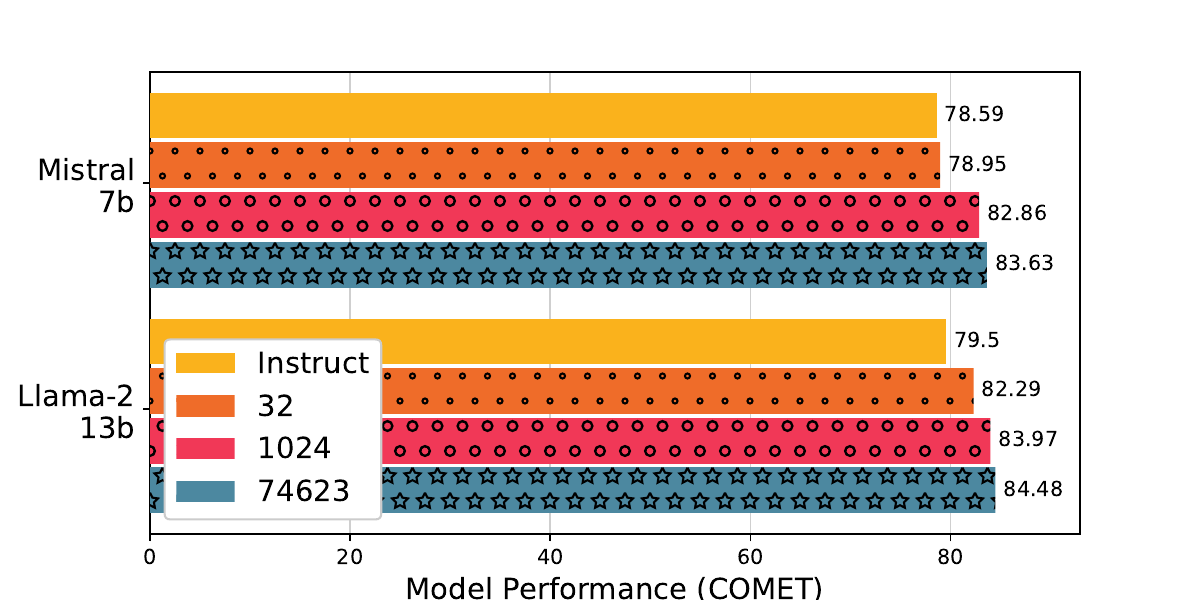}
    \caption{Performance comparison between instruction-tuned baselines and fine-tuned models with different training data sizes. ``Instruct'' refers to the instruction-tuned baselines, specifically \href{https://huggingface.co/mistralai/Mistral-7B-Instruct-v0.1}{Mistral-7B-Instruct-v0.1} and \href{https://huggingface.co/meta-llama/Llama-2-13b-chat-hf}{Llama-2-13b-chat}. "32/1024/74623" represents models fine-tuned on 32, 1024, and 74623 examples, using pre-trained only models: \href{https://huggingface.co/mistralai/Mistral-7B-v0.1}{Mistral-7B-v0.1} and \href{https://huggingface.co/meta-llama/Llama-2-13b}{Llama-2-13b}.}\vspace{5ex}
    \label{fig:more_llms_grouped_bar_plot}
\end{figure*}

While we primarily report the results with Llama-2 7B in our experiments, we hypothesize that state-of-the-art LLMs are largely homogeneous in terms of language distribution and inherent translation capability making our findings applicable to other LLMs. To support this hypothesis, we conduct fine-tuning experiments with Mistral 7B and Llama-2 13B using varying data sizes: 32, 1024, and 70K. As shown in~\Cref{fig:more_llms_grouped_bar_plot}, the general trend is quite similar to the Llama-2 7B case: fine-tuning with 32 examples results in competitive performance, matching or surpassing general-purpose instruction-tuned models. Furthermore, increasing the number of training examples leads to diminishing returns.

\section{Model Performance with Varying Training Directions}
\label{appendix:sec:model_performance_vary_training_directions}
\begin{figure*}[ht]
    \centering
    \includegraphics[width=\linewidth]{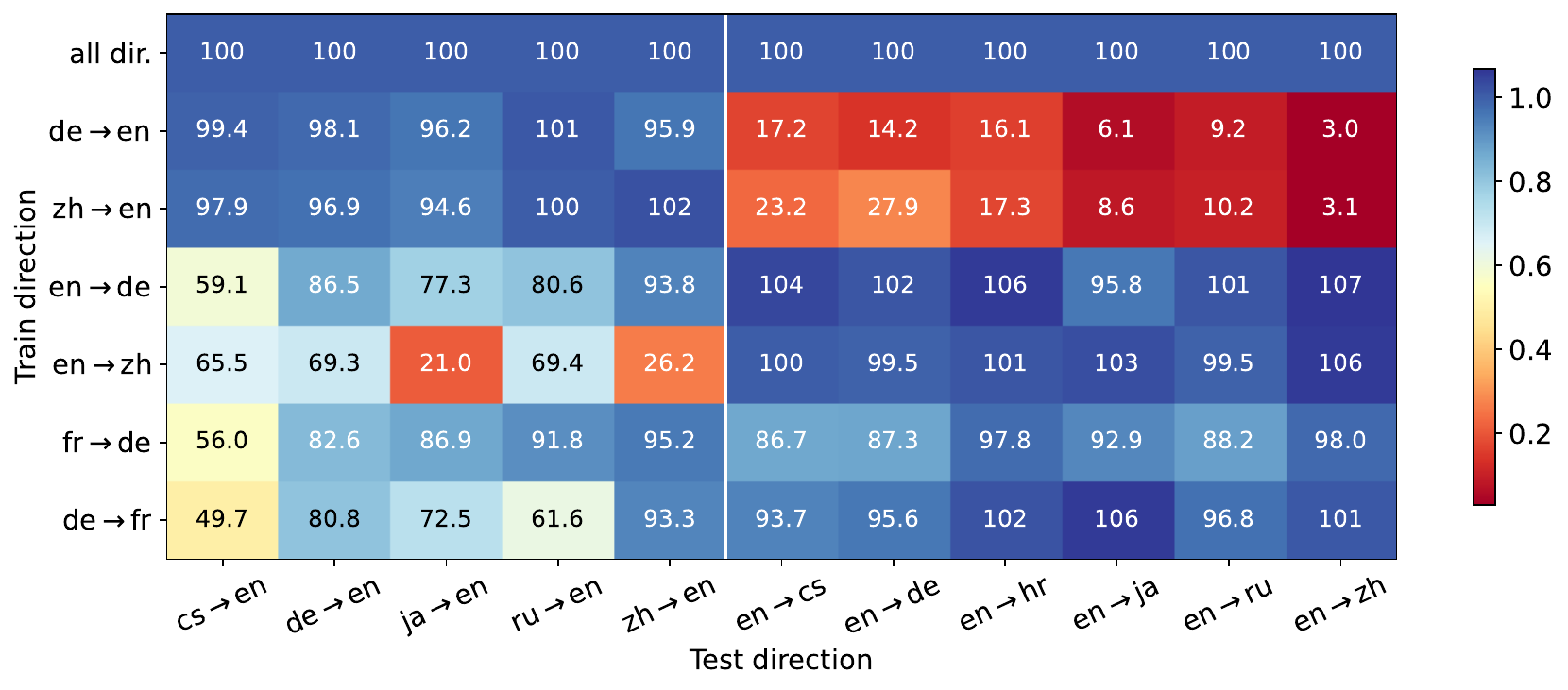}
    \caption{Model performance (\%) in BLEU score resulted from varying combinations of train and test translation directions. The scores are normalized according to Llama-2 fine-tuned on all 10 training directions.}\vspace{5ex}
    \label{fig:single_direction_bleu}
\end{figure*}

Figure~\ref{fig:single_direction_bleu} shows normalized BLEU scores for different combinations of train and test translation directions. Similar to the COMET scores in Figure~\ref{fig:single_direction}, we observe that when training the model on a single direction, its translation ability across other non-targeted directions is also elicited to a certain degree. It is worth noting that when the training direction is X$\rightarrow$en, the performance on directions en$\rightarrow$X is significantly worse than training on all directions. 

\begin{figure*}[ht]
    \centering
    \includegraphics[width=\linewidth]{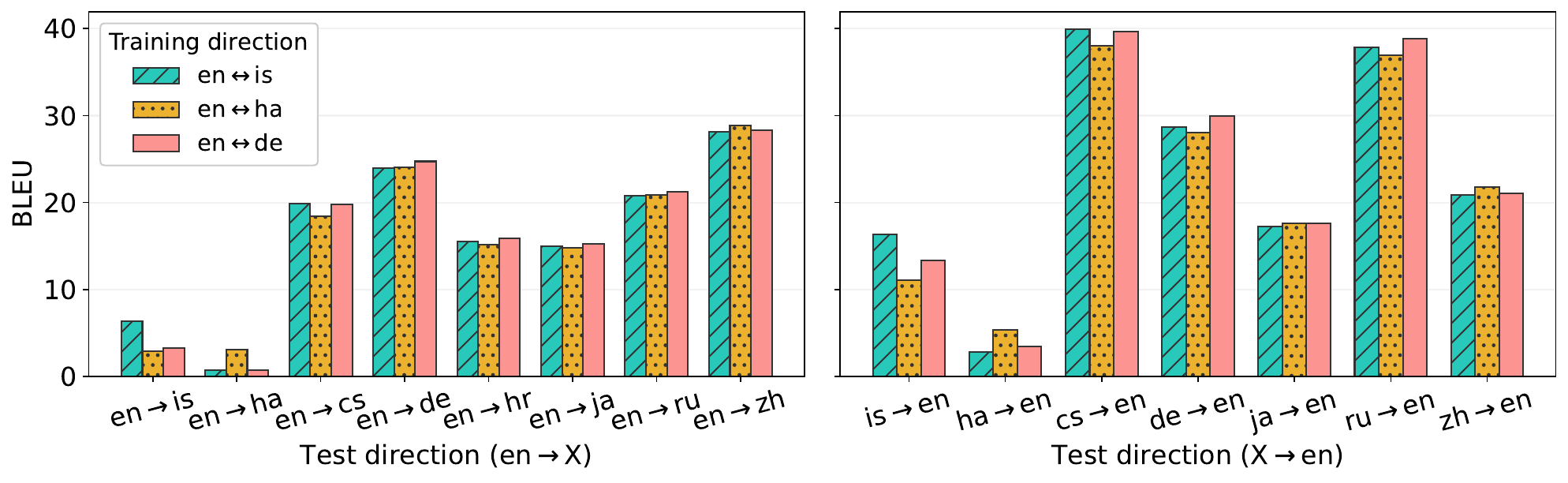}
    \caption{Model performance evaluated across 15 translation directions.  While models trained on \textit{unseen} languages (en$\leftrightarrow$is, en$\leftrightarrow$ha) exhibit moderate improvements in translating these languages, they demonstrate accurate translations from and to \textit{seen} languages.}
    \label{appendix:fig:sft_unseen_direction_bleu}
\end{figure*}

\begin{figure*}[ht]
    \centering
    \includegraphics[width=\linewidth]{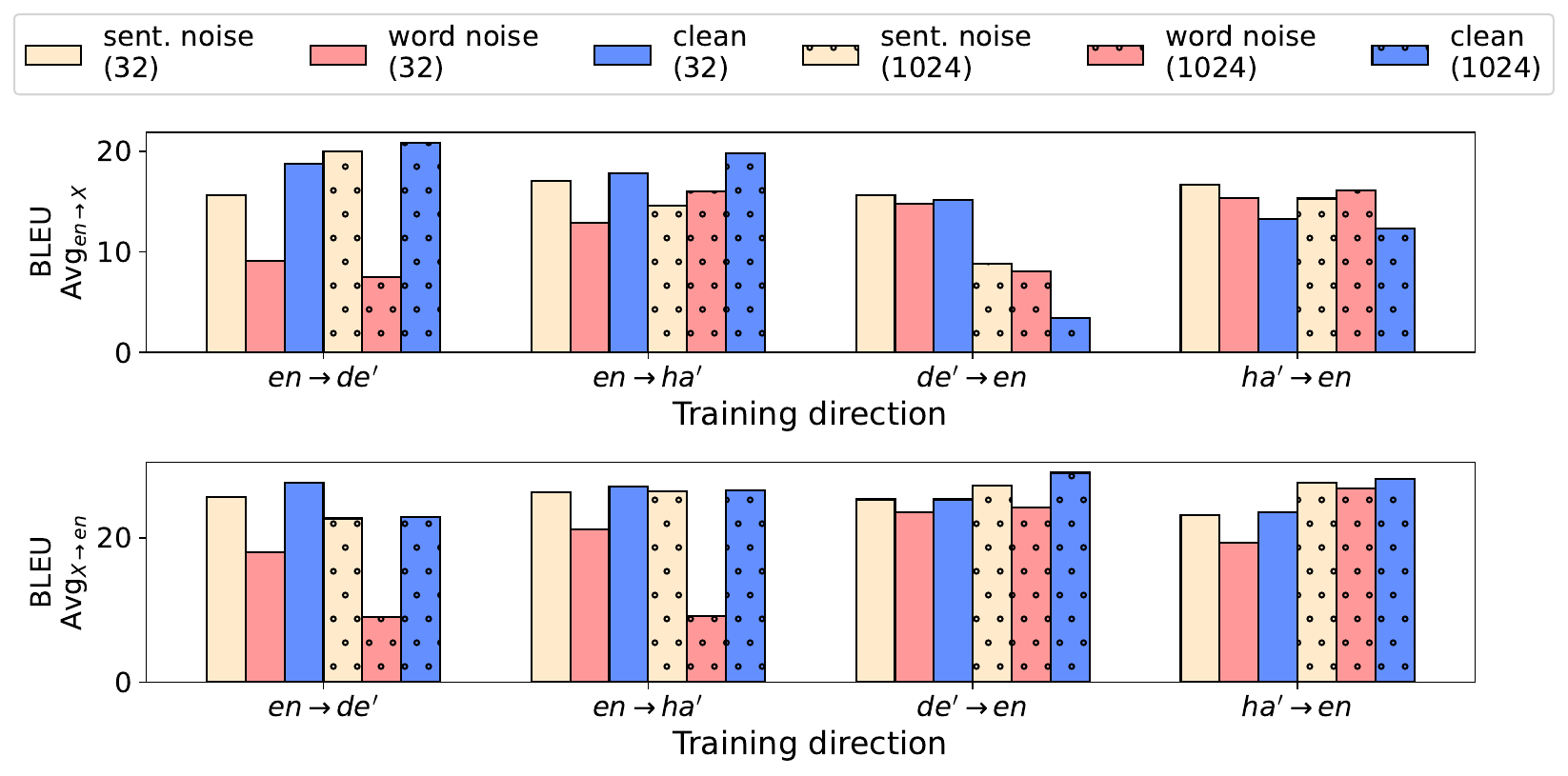}
    \caption{Model performance in BLEU score varying training sizes, directions, and noise types. Top (Bottom): score averaged across all en$\rightarrow$X (X$\rightarrow$en) test directions. Training sizes considered are 32 and 1024.}
    \label{appendix:fig:sft_noise_bleu}
\end{figure*}

\section{Combined Effect of Training Size and Direction}
\label{appendix:sec:model_performance_direction_sep}
\begin{figure*}[ht]
    \centering
    \includegraphics[width=0.49\textwidth]{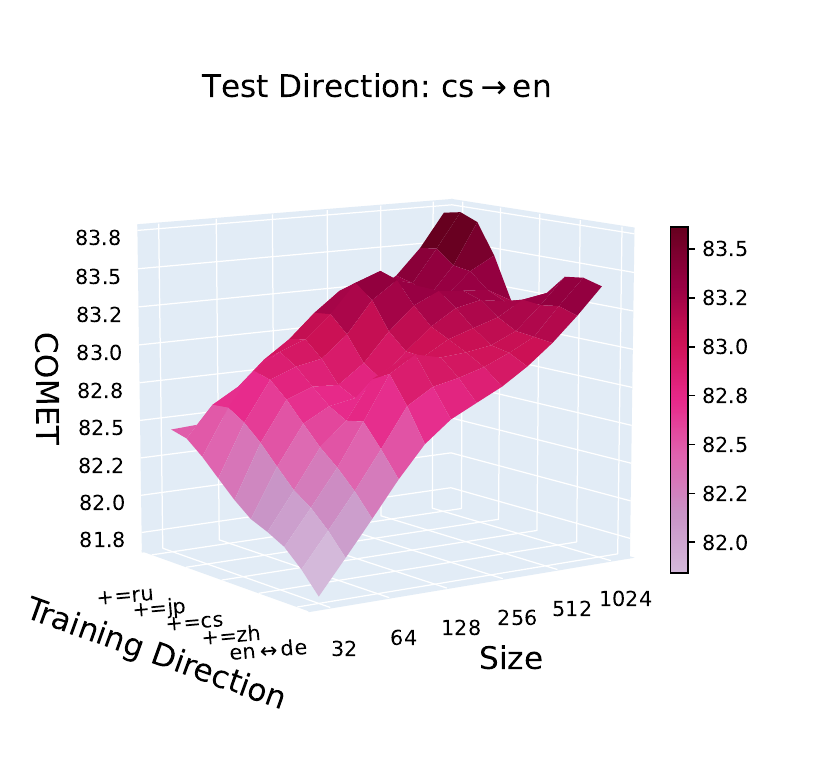}
    \includegraphics[width=0.49\textwidth]{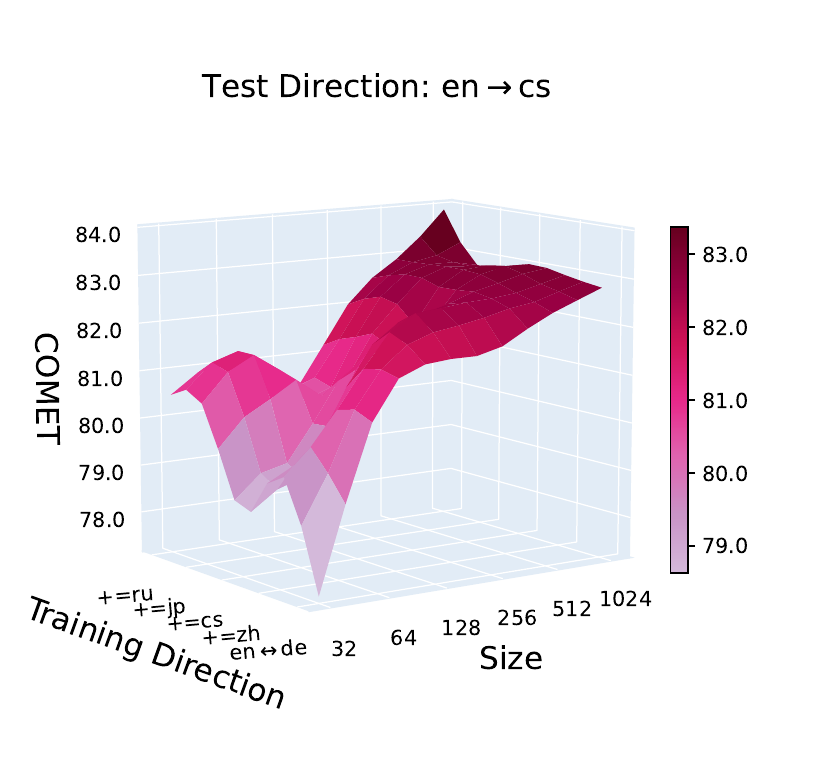}
    \includegraphics[width=0.49\textwidth]{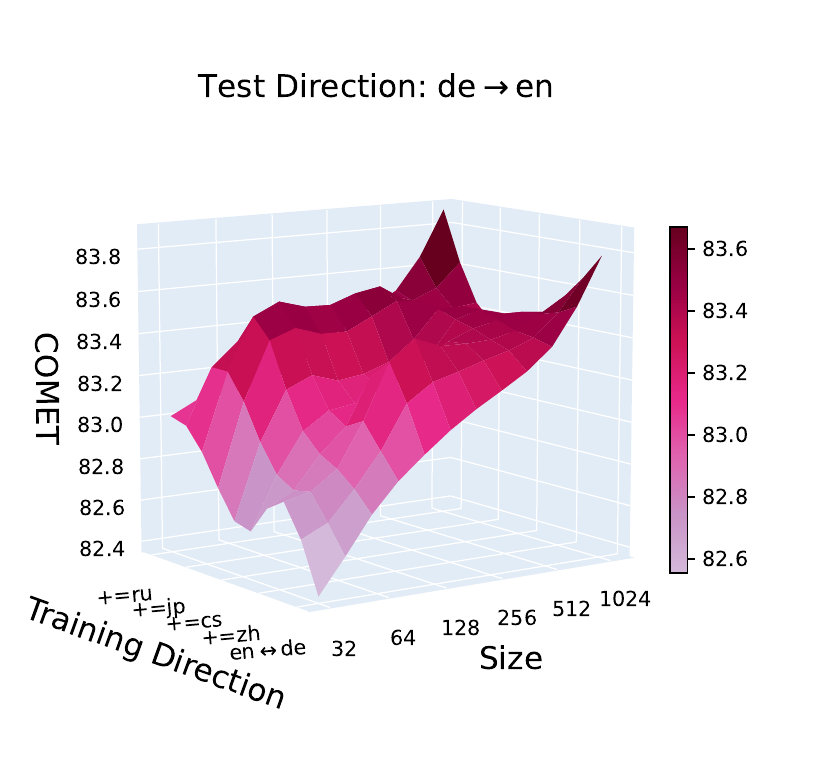}
    \includegraphics[width=0.49\textwidth]{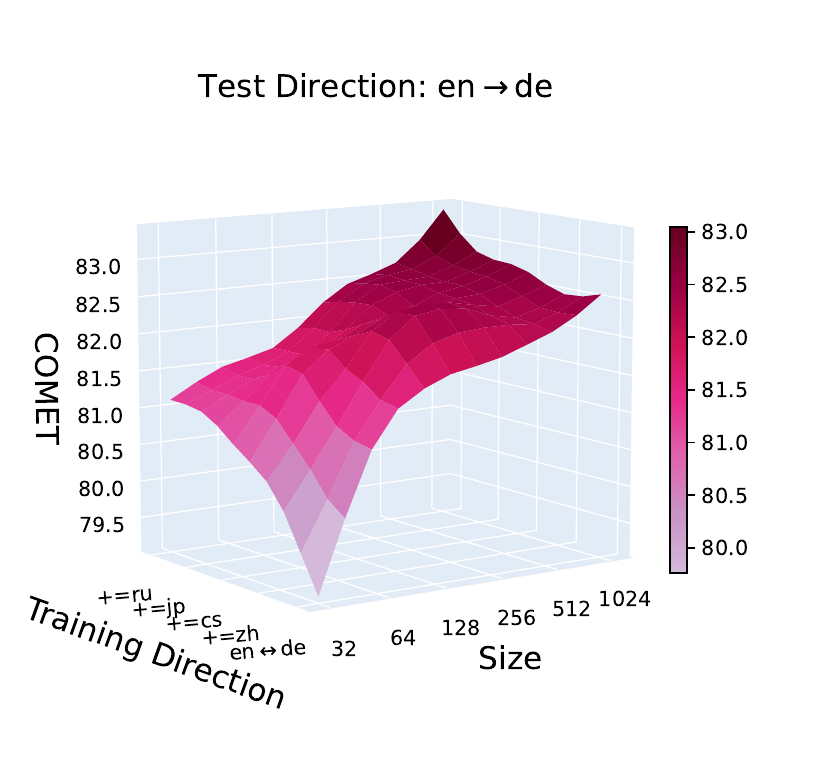}
    \includegraphics[width=0.49\textwidth]{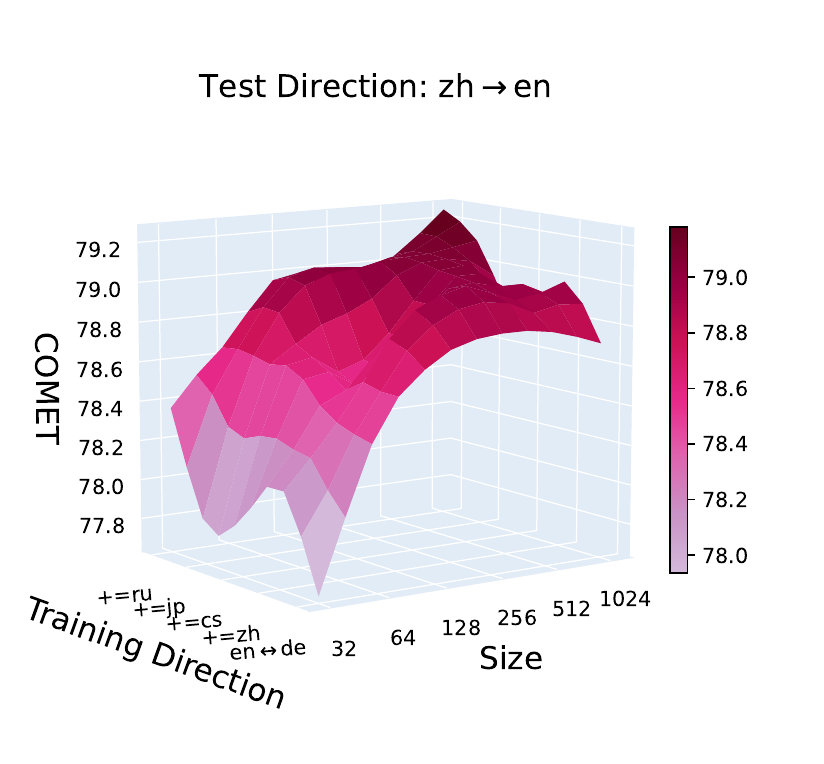}
    \includegraphics[width=0.49\textwidth]{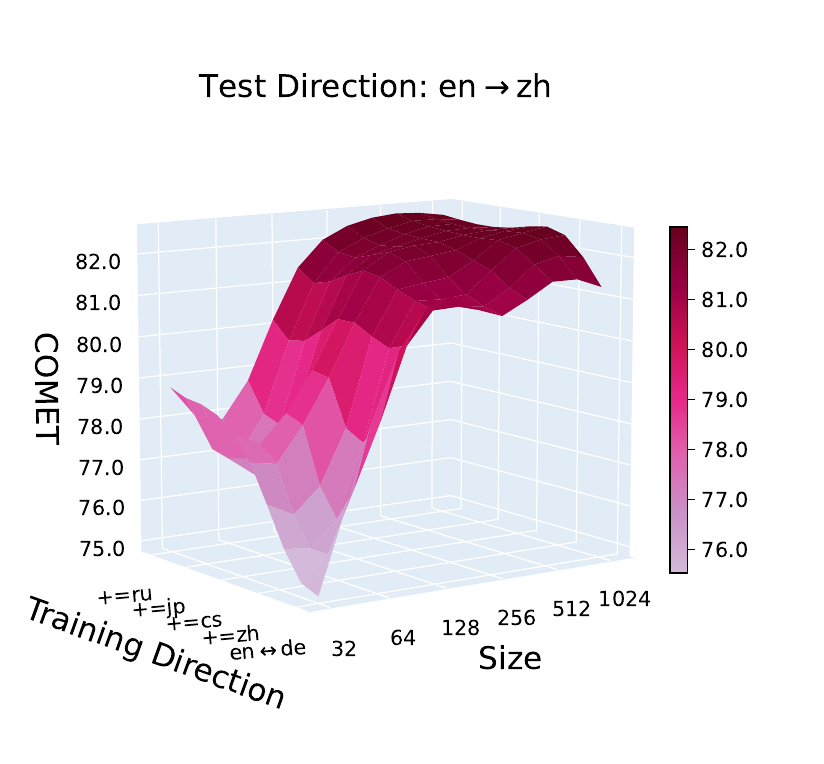}
    \caption{Model performance (in COMET) on individual directions for models trained with varying data sizes and directions. Both factors positively impact performance. +=: training directions added on top of previous directions; two directions (from and to English) at a time. For example, ``+=ru'' covers 10 directions: en $\leftrightarrow$ \{de, zh, cs, jp, ru\}.}
    \label{fig:direction_3d_sep1}
\end{figure*}

\begin{figure*}[ht]
    \centering
    \includegraphics[width=0.49\textwidth]{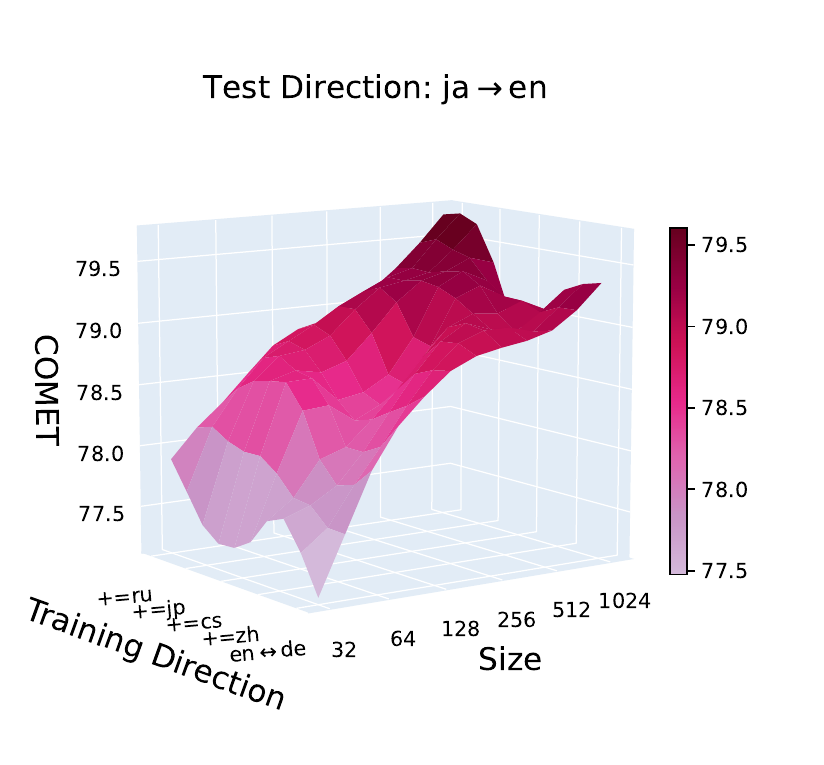}
    \includegraphics[width=0.49\textwidth]{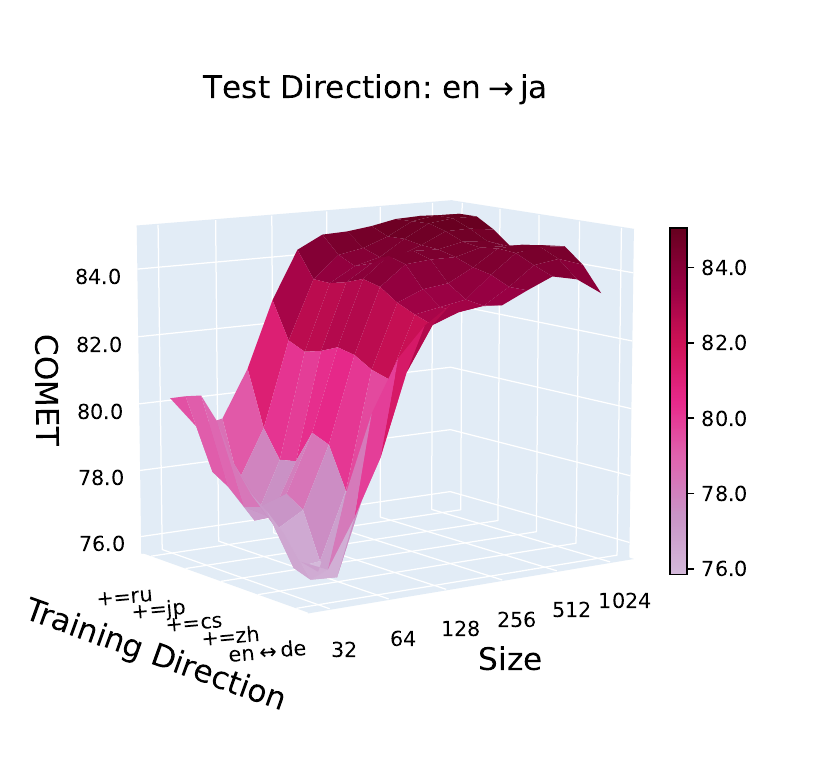}
    \includegraphics[width=0.49\textwidth]{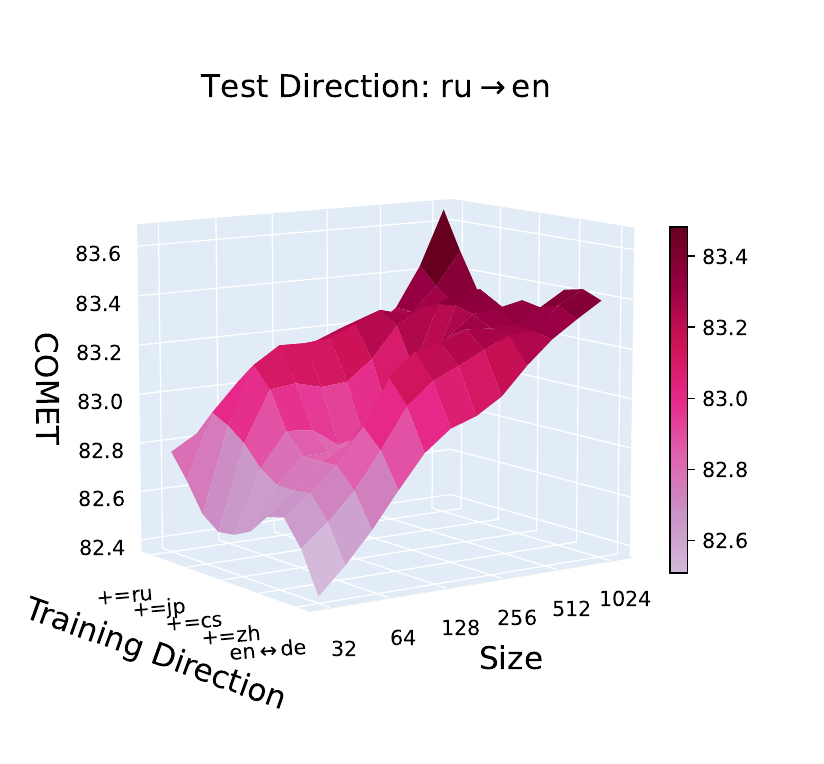}
    \includegraphics[width=0.49\textwidth]{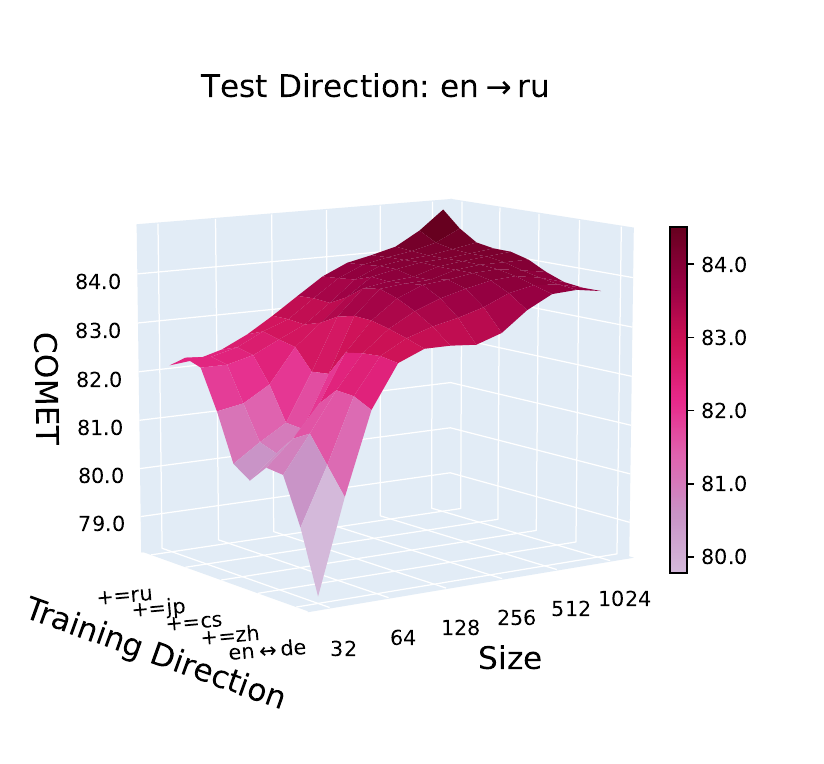}
    \includegraphics[width=0.49\textwidth]{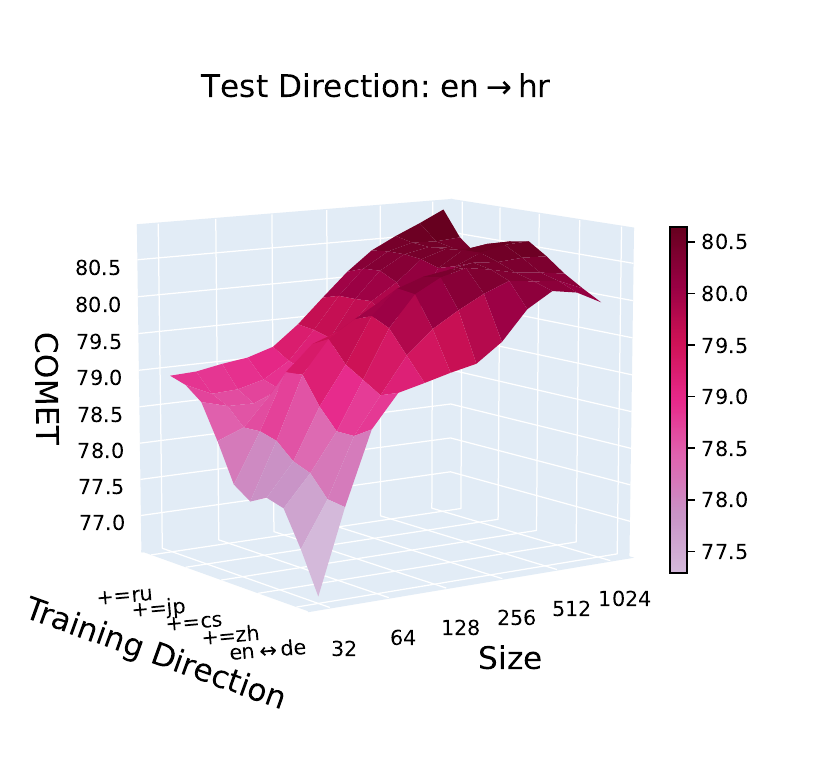}
    \caption{Model performance (in COMET) on individual directions for models trained with varying data sizes and directions. Both factors positively impact performance. +=: training directions added on top of previous directions; two directions (from and to English) at a time. For example, ``+=ru'' covers 10 directions: en $\leftrightarrow$ \{de, zh, cs, jp, ru\}.}
    \label{fig:direction_3d_sep2}
\end{figure*}

\Cref{fig:direction_3d_sep1} illustrates the model performance across varying training sizes and translation directions, evaluated on en$\rightarrow${cs, de, zh}. Similarly, \Cref{fig:direction_3d_sep2} presents the results on en$\rightarrow${cs, de, zh}, and en$\rightarrow$hr. Consistently across all plots, we observe a positive impact on performance with an increasing number of training directions, particularly with smaller training sizes.

\section{Model Performance with Unseen Languages}
\label{appendix:sec:model_performance_unseen_sep}
In Figure~\ref{appendix:fig:sft_unseen_direction_bleu}, we find similar patterns as the COMET score, where fine-tuning on unseen languages can elicit the model's ability to translate from and to all seen languages. However, the translation performance on unseen languages themselves remains subpar, suggesting that SFT primarily reveals the knowledge LLMs have possessed during pre-training.

\section{Model Performance with Noisy Data}
\label{appendix:sec:model_performance_noisy_sep}
Figure~\ref{appendix:fig:sft_noise_bleu} shows the BLEU score of different translation directions with two noise types. We can find that models are more sensitive to word-level noise than sentence-level noise. Also, the performance degradation is more noticeable when injecting noise into the source translation side. In comparison to the results of size 1024, using 32 training examples still achieves comparable or even better performance in the noisy condition.

\section{Technical Details}
\subsection{Datasets}
\label{appendix:sec:dataset_info}
Our parallel data is derived from the development and test sets of WMT17 through WMT22. Detailed dataset statistics are available in~\Cref{appendix:tab:data-statistics}. For most experiments, we use the test sets from WMT17 to WMT20 for training. The test set from WMT22 is used specifically for testing. An exception is noted in Section~\cref{sec:exp_unseen_lanuages}, where models are trained using the en$\leftrightarrow$ha and en$\leftrightarrow$is language pairs from WMT21's development set. Subsequently, these models are evaluated using the corresponding test sets from WMT21.

\subsection{Translation instructions}
\label{appendix:sec:alpaca-template-and-instructions}
The collection of translation instruction templates used in this work can be found in \Cref{appendix:tab:instruction_collection}.

\subsection{Evaluation packages}
\label{sec:comet-bleu-details}
To obtain COMET scores, we use \texttt{Unbabel/wmt22-comet-da}\footnote{\url{https://github.com/Unbabel/COMET}} and for BLEU scores, we use sacreBLEU\footnote{\url{https://github.com/mjpost/sacrebleu}} \citep{post-2018-call}. The signature from the sacreBLEU package is \texttt{nrefs:1, case:mixed, eff:no, tok:13a, smooth:exp, version:2.0.0} for all language pairs, except for tokenization for \texttt{en}$\rightarrow$\texttt{zh} and \texttt{en}$\rightarrow$\texttt{jp}, where we use \texttt{tok:zh} and \texttt{tok:jp-mecab}, respectively.

\begin{table*}[h]
\centering\small
\begin{tabular}{lccccccc}
\toprule
\multicolumn{1}{c}{\multirow{2}{*}{Direction}} & \multicolumn{5}{c}{Training} & \multicolumn{1}{c}{Validation$\color{rred}^*$} & \multicolumn{1}{c}{Test} \\ \cmidrule(lr){2-6} \cmidrule(lr){7-7} \cmidrule(lr){8-8}
               & WMT17 & WMT18 & WMT19 & WMT20 & WMT21dev & WMT21 & WMT22\\
\midrule
en-cs          & 3005  & 2983  & 1997  & 1418 & 0 & 1002 & 2037 \\ 
en-de          & 3004  & 2998  & 1997  & 1418 & 0 & 1002 & 2037 \\
en-hr          & 0     & 0     & 0     & 0     & 0 & 0 & 1671 \\
en-ja          & 0     & 0     & 0     & 1000  & 0 & 0 & 2037 \\
en-ru          & 3001  & 3000  & 1997  & 2002  & 0 & 1002 & 2037 \\
en-zh          & 2001  & 3981  & 1997  & 1418  & 0 & 1002 & 2037 \\
cs-en          & 3005  & 2983  & 0     & 664   & 0 & 1000 & 1448 \\
de-en          & 3004  & 2998  & 2000  & 785   & 0 & 1000 & 1984 \\
ja-en          & 0     & 0     & 0     & 993   & 0 & 1005 & 2008 \\
ru-en          & 3001  & 3000  & 2000  & 991   & 0 & 1000 & 2016 \\
zh-en          & 2001  & 3981  & 2000  & 2000  & 0 & 1948 & 1875 \\\midrule
en-ha          & 0  & 0  & 0  & 0 & 2000 & \textcolor{rred}{1000} & 0 \\
ha-en          & 0  & 0  & 0  & 0 & 2000 & \textcolor{rred}{997} & 0 \\
en-is          & 0  & 0  & 0  & 0 & 2004 & \textcolor{rred}{1000} & 0 \\
is-en          & 0  & 0  & 0  & 0 & 2004 & \textcolor{rred}{1000} & 0 \\\midrule
de-fr          & 0  & 0  & 1701  & 1619 & 0 & $^\otimes$ & 1984 \\
fr-de          & 0  & 0  & 1701  & 1619 & 0 & $^\otimes$ & 2006 \\
\bottomrule
\end{tabular}
\caption{Data statistics. $\mathbf{\color{rred}^*}$Generally, WMT21 test is used for validation purposes; exceptions are en$\leftrightarrow$ha and en$\leftrightarrow$is, which are used for testing. $\mathbf{^\otimes}$Although WMT21 includes data for de$\leftrightarrow$fr, these language pairs are excluded from experiments.}
\label{appendix:tab:data-statistics}
\end{table*}

\subsection{Hardware specifications and runtime}
Our experiments are conducted on a computing node with either 8 NVIDIA A100-40GB GPUs or 8 H100-80GB GPUs. DeepSpeed\footnote{\url{https://github.com/microsoft/DeepSpeed}} with zero-stage 1 and mixed precision bfloat16 is used for performing SFT. Given the limited dataset size, typically fewer than 1024 samples, each SFT experiment can be completed within a mere 15 minutes using four H100 GPUs. However, given the necessity to evaluate the models across more than ten translation directions, the evaluation process may require up to four hours when performed on a single A100-40GB GPU.

\begin{table*}[h]
    \centering\small
    \begin{tabular}{|p{420pt}|}
    \hline
    \rowcolor{gray!20} \textbf{Instruction pool} \\
    \hline
    \texttt{Please provide the \colorbox{blue!30}{[TGT]} translation for the following text} \\
    \hline
    \texttt{Convert the subsequent sentences from \colorbox{orange!30}{[SRC]} into \colorbox{blue!30}{[TGT]}:} \\
    \hline
    \texttt{Render the listed sentences in \colorbox{blue!30}{[TGT]} from their original \colorbox{orange!30}{[SRC]} form:} \\
    \hline
    \texttt{Transform the upcoming sentences from \colorbox{orange!30}{[SRC]} language to \colorbox{blue!30}{[TGT]} language:} \\
    \hline
    \texttt{Translate the given text from \colorbox{orange!30}{[SRC]} to \colorbox{blue!30}{[TGT]}:} \\
    \hline
    \texttt{Turn the following sentences from their \colorbox{orange!30}{[SRC]} version to the \colorbox{blue!30}{[TGT]} version:} \\
    \hline
    \texttt{Adapt the upcoming text from \colorbox{orange!30}{[SRC]} to \colorbox{blue!30}{[TGT]}:} \\
    \hline
    \texttt{Transpose the next sentences from the \colorbox{orange!30}{[SRC]} format to the \colorbox{blue!30}{[TGT]} format.} \\
    \hline
    \texttt{Reinterpret the ensuing text from \colorbox{orange!30}{[SRC]} to \colorbox{blue!30}{[TGT]} language.} \\
    \hline
    \texttt{Modify the forthcoming sentences, converting them from \colorbox{orange!30}{[SRC]} to \colorbox{blue!30}{[TGT]}.} \\
    \hline
    \texttt{What is the meaning of these sentences when translated to \colorbox{blue!30}{[TGT]}?} \\
    \hline
    \texttt{In the context of \colorbox{blue!30}{[TGT]}, what do the upcoming text signify? The text is:} \\
    \hline
    \texttt{How would you express the meaning of the following sentences in \colorbox{blue!30}{[TGT]}?} \\
    \hline
    \texttt{What is the significance of the mentioned sentences in \colorbox{blue!30}{[TGT]}?} \\
    \hline
    \texttt{In \colorbox{blue!30}{[TGT]}, what do the following text convey?} \\
    \hline
    \texttt{When translated to \colorbox{blue!30}{[TGT]}, what message do these sentences carry?} \\
    \hline
    \texttt{What is the intended meaning of the ensuing sentences in \colorbox{blue!30}{[TGT]}?} \\
    \hline
    \texttt{How should the following sentences be comprehended in \colorbox{blue!30}{[TGT]}?} \\
    \hline
    \texttt{In terms of \colorbox{blue!30}{[TGT]}, what do the next sentences imply?} \\
    \hline
    \texttt{Kindly furnish the \colorbox{blue!30}{[TGT]} translation of the subsequent sentences.} \\
    \hline
    \texttt{Could you supply the \colorbox{blue!30}{[TGT]} translation for the upcoming sentences?} \\
    \hline
    \texttt{Please offer the \colorbox{blue!30}{[TGT]} rendition for the following statements.} \\
    \hline
    \texttt{I'd appreciate it if you could present the \colorbox{blue!30}{[TGT]} translation for the following text:} \\
    \hline
    \texttt{Can you deliver the \colorbox{blue!30}{[TGT]} translation for the mentioned sentences?} \\
    \hline
    \texttt{Please share the \colorbox{blue!30}{[TGT]} version of the given sentences.} \\
    \hline
    \texttt{It would be helpful if you could provide the \colorbox{blue!30}{[TGT]} translation of the ensuing sentences.} \\
    \hline
    \texttt{Kindly submit the \colorbox{blue!30}{[TGT]} interpretation for the next sentences.} \\
    \hline
    \texttt{Please make available the \colorbox{blue!30}{[TGT]} translation for the listed sentences.} \\
    \hline
    \texttt{Can you reveal the \colorbox{blue!30}{[TGT]} translation of the forthcoming sentences?} \\
    \hline
    \texttt{Translate from \colorbox{orange!30}{[SRC]} to \colorbox{blue!30}{[TGT]}:} \\
    \hline
    \end{tabular}
    \caption{A collection of 31 translation prompts. Each instruction is randomly selected to form a training sample. At inference time, the first instruction is always selected. The placeholders \colorbox{orange!30}{[SRC]} and \colorbox{blue!30}{[TGT]} represent the source and target languages, respectively, and will be replaced with the appropriate languages depending on the specific example at hand.}
    \label{appendix:tab:instruction_collection}
\end{table*}

\end{document}